\documentclass{article}

\usepackage{arxiv}

\usepackage{booktabs}       
\usepackage{nicefrac}       
\usepackage{microtype}      
\usepackage{lipsum}
\usepackage{fancyhdr}       
\usepackage{graphicx}       
\graphicspath{{./img/}}     

\usepackage[utf8]{inputenc}
\usepackage[LAE,T1]{fontenc}
\usepackage{arabtex}
\usepackage{utf8}
\setcode{utf8}
\usepackage[english]{babel}

\usepackage{amsmath,amsfonts}
\usepackage{algorithmic}
\usepackage{algorithm}
\usepackage{array}
\usepackage[caption=false,font=normalsize,labelfont=sf,textfont=sf]{subfig}
\usepackage{textcomp}
\usepackage{stfloats}
\usepackage{url}
\usepackage{verbatim}
\usepackage{wrapfig}
\usepackage{float}
\usepackage{cite}
\usepackage{listings}
\usepackage{fancyvrb}
\usepackage{framed}
\usepackage[listings,skins]{tcolorbox}
\usepackage[skipbelow=\topskip,skipabove=\topskip]{mdframed}
\mdfsetup{roundcorner=0}

\usepackage{multirow}
\usepackage{longtable}

\lstdefinestyle{mycoding}{
  backgroundcolor=\color{lightgray}, commentstyle=\color{purple},
  keywordstyle=\color{magenta},
  numberstyle=\tiny\color{darkgray},
  stringstyle=\color{violet},
  basicstyle=\ttfamily\footnotesize,
  breakatwhitespace=false,         
  breaklines=true,                 
  captionpos=b,                    
  keepspaces=true,                 
  numbers=none,                    
  numbersep=5pt,                  
  showspaces=false,                
  showstringspaces=false,
  showtabs=false,                  
  tabsize=2  
}
\usepackage{xltabular}
\usepackage{tabularx}

\usepackage{enumitem}

\usepackage[hidelinks,pagebackref=true]{hyperref}

\title{Code-Switched Urdu ASR for Noisy Telephonic Environment using Data Centric Approach with Hybrid HMM and CNN-TDNN}

\author{ \href{https://orcid.org/0009-0001-5084-6122}{\includegraphics[scale=0.06]{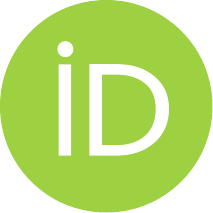}\hspace{1mm}Muhammad Danyal "Sage" Khan}\thanks{www.thesageforce.com} \\
	Department of Cyber Security \\
	Pakistan Navy Engineering College \\
	National University of Science and Technology \\
        Karachi, Pakistan \\
	\texttt{sagekhanofficial@gmail.com} \\
	\And
	\href{https://www.linkedin.com/in/alirahim/}{\includegraphics[scale=0.06]{orcid.pdf}\hspace{1mm}Raheem Ali} \\
	Department of Computer Science\\
	Bahria University Karachi Campus\\
	Karachi, Pakistan \\
	\texttt{rahim.ali@me.com} \\
        \And
	\href{https://www.linkedin.com/in/arshad-aziz-ph-d-2b0a99172/}{\includegraphics[scale=0.06]{orcid.pdf}\hspace{1mm}Dr. Arshad Aziz} \\
	Department of Cyber Security \\
	Pakistan Navy Engineering College \\
	National University of Science and Technology \\
        Karachi, Pakistan \\
	\texttt{dr\_arshadaziz@yahoo.com} \\
}

\renewcommand{\shorttitle}{\textit{Code-Switched} Urdu ASR for Noisy Telephonic Environment}

\begin{document}

\maketitle

\begin{abstract}
Call Centers are an effective medium for businesses, organizations and even government institutions to interact with their clients despite availability of mobile apps and websites. These centers have huge amount of audio data which can be used for achieving valuable business insights. This requires transcription of phone calls which is manually tedious task to perform. An effective Automated Speech Recognition system can accurately transcribe these calls, making it easier to search through call history for specific context and content with very less time and manual effort. With text transcription available calls can be automatically monitored, improving QoS through keyword search and sentiment analysis. ASR for Call Center requires more robustness as telephonic environment are generally noisy. Moreover, there are many low-resourced languages that are on verge of extinction which can be preserved with help of Automatic Speech Recognition Technology. Urdu is the $10^{th}$ most widely spoken language in the world, with 231,295,440 worldwide still remains a resource constrained language in ASR. Regional call-center conversations operate in local language, with a mix of English numbers and technical terms generally causing a "code-switching" problem. Hence, this paper describes an implementation framework of a resource efficient Automatic Speech Recognition/ Speech to Text System in a noisy call-center environment using Chain Hybrid HMM and CNN-TDNN for Code-Switched Urdu Language. Using Hybrid HMM-DNN approach allowed us to utilize the advantages of Neural Network with less labelled data. Adding CNN with TDNN has shown to work better in noisy environment due to CNN's additional frequency dimension which captures extra information from noisy speech, thus improving accuracy. We collected data from various open sources and labelled some of the unlabelled data after analysing its general context and content from Urdu language as well as from commonly used words from other languages, primarily English and were able to achieve WER of 5.2\% with noisy as well as clean environment in isolated words or numbers as well as in continuous spontaneous speech.
\end{abstract}


\section{Introduction}
With the advent of telecommunications and the internet, the world has become a global village. Nowadays, almost everyone owns a cellphone, making telecommunication the most effective means of communication available to humanity. Call Centers are one of the most popular communication channels used by government institutions, businesses, and other organisations as a medium of interaction between the firm and the consumers, quickly becoming a powerful service delivery  approach and an important operational component of many businesses around the world \cite{chuchual_inbound_2010}. Companies such as Douglas, Depot, H\&M, Zara, Kaufhof, Commerzbank Germany \cite{commerzbank_commerzbank_2021} and Esprit also announced that they would be closing numerous stores permanently in 2021 and relying increasingly on online retailing with acceleration in the global digitization process. The number of call centre agents in Germany more than doubled between 2000 and 2012, rising from 220,000 to 520,000. \cite{herzog_callcenter_2017}. 


Local stores are now replaced by E-Commerce stores, local services like banking are replaced by digital services like online banking and brick \& mortar retail is replaced by online shopping thanks to the internet. Companies have various channels to communicate with their customers beyond traditional phones like emails, chats, video conferencing etc which must be handled with highest level of quality to handle customer inquiries for which various Customer Relationship Management (CRM) are in place \cite{glas_einzelhandel_2017}. 

Despite the availability of these channels human agents are still preferred by customers over digital ones which is why Call centers continue to thrive. Given the choice, 83\% of consumers will still prefer interaction with humans agents as they understand customer needs and can address multiple questions at once. Bots, in comparison to humans, are currently incapable of dealing with complicated requests, delivering personalised offers, or human emotions comprehension \cite{forrester_human_2017} making Human Agents for call center an irreplaceable asset. 


Call Centers generate huge amount of Speech data containing insights about customer experiences and expectations which are often under utilized for business purposes. Speech audio data usually is in wave files which are not searchable. Either they require manual search and analysis for specific business related information  or through speech analytics, caller and agent audio conversations can be automatically used to derive business insights and other statistically significant information \cite{liu_research_2012}. 


Speech analytics system uses ASR technology to automatically transcribe the audio conversations into text making it searchable, process and analyze it to provide customer profiled based classification using common characteristics, keywords search \cite{fasel_digitalisierung_2016} etc. Evaluations of call center operations like call center agent assessment for work standard compliance, number of product complaints, product reviews, latest trends, type of problems, comparative sales trend between products \cite{draman_malay_2017} etc aim to successfully and profitably use the insights gained from them to optimize their digital products and central service centers for their end-users, generating further survey data at same time to use it for competitive advantage \cite{nollenburg_visual_2022}.

Call center ASR is an active research area and has been implemented in English \cite{bernstein_recognizing_2000} but their language and acoustic model does not apply to relatively low-resourced South Asian Regional languages. Call centers operating in regional languages use English jargon, numbers and terminologies, which presents a code-switching problem \cite{farooq_enhancing_2020}. Urdu, although spoken by 231 Million people around the world \cite{ethnologue_urdu_nodate}, is still a low resourced language and the problem is amplified in a noisy telephonic code-switching scenario, which is a major problem in an Call center operating in Urdu-speaking region.

Thus we proposed a Data Centric Approach to building a Code-switched Urdu ASR for telephonic Environment in a resource constrained environment using Hybrid HMM-DNN which allowed us to train ASR with up to 95\% accuracy with just 10 hours of labelled data. Aim of the research is to develop an effective code-switched Roman Urdu language model with easily deployable Speech to Text interface and implement it to give an accurate speech to text in a call center environment with good WER and SER, following objectives must be completed:


\begin{itemize}
    \item Building Large Vocabulary ASR for code switched Urdu Language using state of the art techniques after careful comparison. 
    \item Selection of resource effective method (in terms of time, money, computational power, Hardware and HR) for ASR implementation.
    \item Collection of code-switched Urdu telephonic data (covering Large Vocabulary).
    \item Building an effective code-switched Urdu language model using Roman Urdu script.   
    \item Selection of suitable ML implementation model and ASR Training method to achieve Word Error Rate (WER) of less than 10\% and Sentence Error Rate (SER) of less than 30\% in a noisy telephonic environment.
\end{itemize}


\section{Literature Review}
\label{sec:literature_review}


Building an Automated Speech to Text (ASR) System especially in a call center environment to transcribe massive audio data-sets and extract valuable insights entails converting the audio conversation between the caller and agent into text transcripts with help of statistical or Neural Network trained on big audio dataset, followed by text transcripts analysis using natural language processing to derive useful analytics \cite{kopparapu_non-linguistic_2015, plaza_call_2021}.

There is ample amount of work done in English Language with respect to Automatic Speech Recognition (ASR). All state-of-the-art Training methods ranging from Traditional statistical ASR to Hybrid Hidden Markov Model-Deep Neural Network (HMM-DNN) to modern End to End (E2E) Systems are thoroughly tested on English Language with WERs $<6\%$ \cite{georgescu_performance_2021}. The Speech Recognition problem primarily involves prediction of word based on incoming audio signal which requires a language and acoustic model \cite{backstrom_introduction_2022} which was done statistically from 1970s-early 200s using HMM-GMM but with availability of GPUs and higher processing power, NNs have gained popularity to train ASRs \cite{bell_adaptation_2020}. 

Dahl et al 2012 \cite{dahl_context-dependent_2012} first introduced HMM–DNN hybrid approach for English ASR, achieving improvement of Absolute Sentence Accuracy by 16\% in comparison to traditional HMM-GMM legacy system on English data set containing audio with background music, noise, hesitation etc. Similarly, Hu et al 2022 \cite{hu_neural_2022} achieved Word Error Rates (WERs) of 9.9\% on the conversational telephonic audio corpus \cite{linguistic_data_consortium_2000_2002, fiscus_jonathan_g_2003_2007} with model size reduction of 96\%. Furthermore in 2021, Georgescu et al. \cite{georgescu_performance_2021} studied and compared the performance of Hybrid HMM-DNN and Deep Learning based methods trained on English LibriSpeech data-set of 1000 hours, which showed that Hybrid HMM-DNN based systems, particularly Time Delay Neural Network (TDNN) and Convolutional Neural Network - Time Delay Neural Network (CNN-TDNN) outperformed End to End Methods with WER of 3.85\% and 3.87\% respectively with least memory load of 106 MB for loading Neural Network model and storing all the activation's of the network for processing per second of speech. While these works achieved significant results, English models could not be used in Urdu language since every language requires a different acoustic and language model.

Arabic Language is one of the most widely spoken languages in the world and is one of the languages from which Urdu is derived. Alsayadi et al 2021 \cite{alsayadi_arabic_2021} trained Models with Traditional and E2E methods with 1200 hours of Arabic Speech Corpus \cite{alhanai_development_2016} achieving WER of 33.72\% on conventional ASR and 28.5\% on CNN-LSTM. This was not applicable in our case as the WER is too high and training this model required 1000s of hours of high-quality labelled code-switched Urdu data-set which was not available. 




The majority of work in the field of speech recognition in South Asian Languages has been done on Hindi, which is very similar to spoken Urdu \cite{farooq_improving_2019} \cite{dash_automatic_2018}. Isolated word recognition \cite{patil_automatic_2016, lakshmi_sri_kaldi_2020}, connected digit recognition \cite{a_n_mishra_robust_2011}, statistical pattern classification \cite{aggarwal_using_2011}, online speech to text engine \cite{b_venkataraman_sopc-based_2006} and large vocabulary speech recognition systems have been developed in Hindi \cite{kumar_large-vocabulary_2004, k_v_s_parsad_and_s_m_virk_computational_2012, tanveer_continuous_2019} with accuracies up to 95\%. While Lakshmi et al 2020 \cite{lakshmi_sri_kaldi_2020} achieved 2\% WER,  Hindi Language Model, even in Roman Hindi Script, cannot be used to implement in Code-switched Urdu telephonic environment due to differences in the linguistic foundation, pronunciation, vocabulary, written scripts, etc \cite{k_v_s_parsad_and_s_m_virk_computational_2012} e.g. Phones like \textit{"/Z/"} \textit{"/kh/"} \textit{"/gh/"} are pronounced differently in Hindi and in Urdu which is why the word \textit{"Zeera"} in Urdu is pronounced \textit{"Jeera"} in Hindi (in writing and in pronunciation both), thus resulting in low accuracy for our scenario.

To the best of our knowledge, no model or large labelled data-set was available for implementation of ASR in Code-Switched Urdu for telephonic call centre environment, so an ASR with effective language model and acoustic model comprising of individual words and spontaneous speech in noisy as well as clean environment had to be built from scratch. Due to data privacy concerns, domain-specific telephonic data is scarce, and telephone calls frequently have highly distorted signals due to transmission impairments \cite{tits_sector_effect_1993, chunwijitra_improving_2021}, as is the case for Urdu language as well. 

Raza \cite{raza_rapid_2018} achieved WER of 24.14\% on 1207 hours of Urdu audio with vocabulary of 5000 words which was too high for our scenario. Farooq et al 2019 \cite{farooq_improving_2019} developed an LVCSR using 300 hours of read-out speech data, in both indoor and outdoor environments, with a vocabulary size of 199,000 words from 1671 Urdu and Punjabi speakers using GMM-HMM, TDNN, LSTM, and Bidirectional-LSTM for Acoustic Modelling and Recurrent Neural Network Language Model (RNNLM) for re-scoring giving WER of 13.5\% which was too high and did not cater for code-switching with English words. Their subsequent work \cite{farooq_enhancing_2020} in 2020 achieved a Word Error Rate (WER) 26.95\% with 25 hours of their own Urdu speech corpus for enhancement of read-Urdu-speech LVCSR to recognize code-switched speech using the HMM-DNN modeling technique without any prior GMM-HMM training and alignments while also deploying various techniques to improve language models using monolingual data. This work tried to cater for code-switching but it is not applicable call center scenario due to high WER which is likely to increase in noisy telephonic environment.

Naeem et al 2020 \cite{naeem_subspace_2020} developed SGMM based ASR model for continuous speech with their own 100 hours mono-channel audio data-set with sampling rate of 16000Hz achieving WER of 9.7\%. Unfortunately this dataset is not openly available and is not tested on code-switched noisy telephonic environment. Experiments on the PRUS data-set \cite{zia_pronouncur_2018} comprising of 708 code-switched Urdu sentences by 7 speakers were conducted \cite{qureshi_urdu_2021} with GMM based triphone acoustic models which gave overall recognition accuracy values 78.2\% for 100 words making the model inapplicable for noisy telephonic environment as WER is too high and the vocabulary is very limited. 




Compared to lattice based MMI model, LF-MMI objective function for Neural Network Training gives better performance on various data-sets with HMM-DNN and E2E ASR training frameworks \cite{daniel_povey_kaldi_nodate} \cite{povey_purely_2016}. CNN-TDNN with LF-MMI Objective function has been found to perform better on noisy data-sets \cite{noauthor_tdnn_nodate}, which is supported by Kreyssig et al 2018 \cite{kreyssig_improved_2018}, Abdul Hamid et al 2014 \cite{abdel-hamid_convolutional_2014}, Zorila et al 2019 \cite{zorila_investigation_2019}, Biswas et al 2019 \cite{biswas_semi-supervised_2019}, Georgescu et al 2019 \cite{georgescu_kaldi-based_2019} and Georgescu et al 2021 \cite{georgescu_performance_2021}. Hence for us, Hybrid HMM with CNN-TDNN a good ASR training approach for noisy telephonic dataset and to the best of our knowledge, this approach has not been taken before for telephonic code switched Urdu ASR. This architecture has CNN Layers followed by TDNN Layers. TDNN works well in Speech Processing compared to other Neural Networks as they can be easily tuned in comparison to LSTMs, and gives 11.4\% WER which is better than than the baseline TDNN's 12.1\% WER on the Switchboard \cite{godfrey_switchboard_1992} part of Eval2000 English conversational telephonic data-set. It also works much better on streaming data compared to other neural networks because of its pooling and sub-sampling feature \cite{daniel_povey_kaldi_nodate}. Adding CNN before TDNN layers provide can help the model to learn temporal and frequency domain information which helps it to learn speech from noisy telephonic data, allowing better performance on noisy data-set than pure TDNNs.

\begin{figure}[htb]
    \centering
    \includegraphics[width=0.8\textwidth]{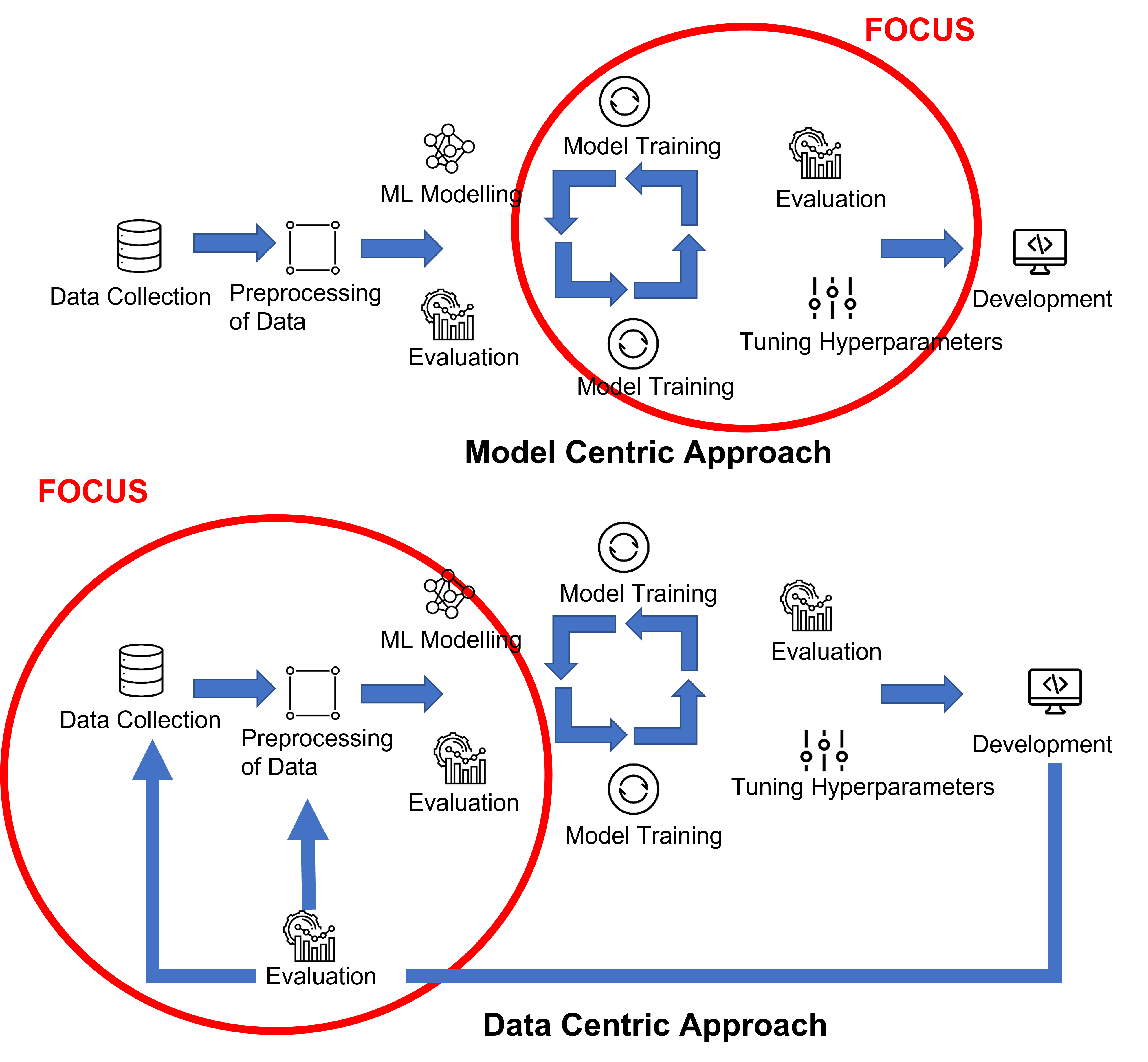}
    \caption{Data-Centric Vs Model-Centric Approach}
    \label{fig:datacent-vs-modelcent}
\end{figure}

Another problem that persists is the model-centric approach for AI implementation, which applies to ASRs as well. The majority of AI applications are currently model-centric, a possible reason being that AI industry pays close attention to academic research on models. Since it is difficult to create large data-sets which can become widely accepted standards, more than 90\% of AI research papers are model-centric \cite{deeplearningai_data-centric_2021}. The Code becomes the major focus and data-set is frequently overlooked which is why data collection and pre-processing is considered as a one-time event. Accenture \cite{noauthor_scaling_nodate} informs that about 85\% of projects are still proof of concept \cite{noauthor_data-centric_nodate} which are not brought into production, yet. There is another approach called "Data-Centric approach" which includes focusing on Data Preparation primarily, leading to better input for the machine learning process because ASR model quality cannot out perform its data quality. The focus is to make a good quality Data-set instead of only focusing on code or tweaking the model as shown in Figure \ref{fig:datacent-vs-modelcent}.

The reason why AI applications have not been deployed in production environment is because there is no one size fits all AI solution and usually there are limited points of data collection which gives less data for AI to train on. There are various types of research claiming things that are not actually deployed in a real-world scenario e.g. in 2019, it was published \cite{ardila_end--end_2019} that certain solutions were developed to accurately detect tumours at an earlier stage than trained radiologists are able to diagnose. However this model is not in every hospital yet because of the substantial gap between the proof of concept and hospital production software. Same is the case with a full fledged ASR application in call centers which is why a practical, robust and scalable implementation approach for application of ASR in a call center environment is essentially required.

It is evident from our literature review that there are some models available in Urdu language but they can not work in noisy telephonic environment, especially in code-switching scenario which is the main issue faced in Call Centers. In this regard we need:

\begin{itemize}
    \item Robust ASR training approach and neural network based model with suitable objective function to accurately train our data
    \item Integration of Data Centric Approach to ASR training
    \item Telephonic code-switched Urdu audio data-set (Mono channel 16000Hz) with Roman Urdu Transcript for training and testing, which would cater for Isolated Digits, Isolated words, Read Speech, Spontaneous Speech (Words and Digits), Clean Audio and Noisy or Telephonic Audio.
    \item Achieving WER $<$ 10\% and SER $<$ 30\% in above scenarios. 
\end{itemize}


\section{Our Methodology}

We proposed a \textit{"Data Centric Approach with Hybrid HMM-DNN ASR Training method using Convolutional Neural Network-Time Delay Neural Network (CNN-TDNN) with Lattice Free Maximum Mutual Information (LF-MMI) objective function for building a Code-switched Urdu ASR"}. The model centric approach has its own set of challenges as explained in Section \ref{sec:literature_review}. 

\textit{"Data-centric Approach"} involves Iterative Collection of Data from all possible sources as explained in subsection \ref{sub:datasources} to improve the performance of ASR as well as the quality and quantity of data and Iterative evaluations for Data Drift Detection. We also have to ensure Quality Data Labelling since labelling in our case was manually done leading to human errors and biases because if there are inconsistencies regarding how the human experts approach particular problem, machines are not likely to detect those inconsistencies. This requires clearly defined Labelling instructions (the process which is always to be improved through out implementation stages), keeping positive control of labelling annotation process and inclusion of Domain knowledge and expertise i.e. looking at the problem not only as AI problem but as a Linguistic problem as well and keeping end-user problems or feedback in mind as well.

Moreover, we use Data Augmentation to increase the number of data points in sample which was done by generating the unseen data that your model has not seen during the training time like our own recordings, removing or adding the noisy observations to cater for maximum possible scenarios, altering speed of speech audio, altering to pitch or tone, altering audio length, adding silences and extra noises using audio editing tools \cite{audacity_linux_nodate} for added variety.

\begin{figure}[h]
    \centering
    \includegraphics[width=0.7\textwidth]{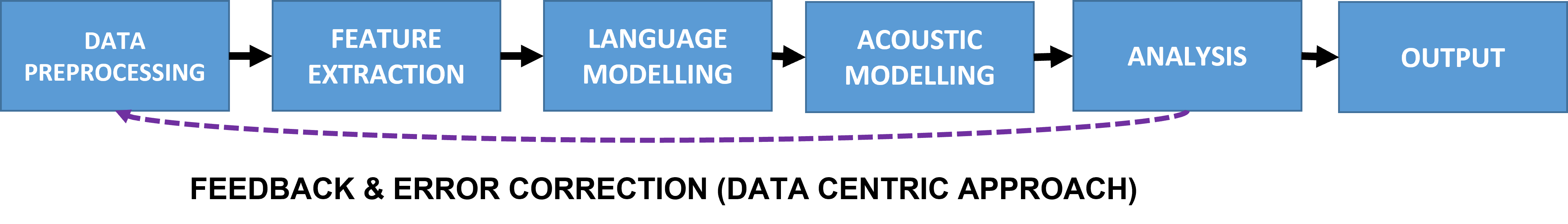}
    \caption{ASR Training Working Pipeline}
    \label{fig:working_pipeline_short}
\end{figure}

Figure \ref{fig:working_pipeline_short} shows the birds eye view of our working pipeline. Every ASR systems requires a data-set in a certain format for feature extraction and modelling which is why Data Pre-processing is the first step. This is the most crucial step because errors in this step will propagate ahead which later becomes very hard to detect. No ASR Model can outperform the quality of its data-set. We used a lot of diversity in our data-set to cater for various scenarios like isolated digits and alphabets, continuous speech etc in clean and noisy backgrounds. After pre-processing, audio features are extracted in a format that is understood by the computer. In order to provide a background understanding of the structure of language, Language model is trained which is then followed by acoustic modelling. 

\begin{figure*}[h]
    \centering
    \includegraphics[width=1.0\textwidth]{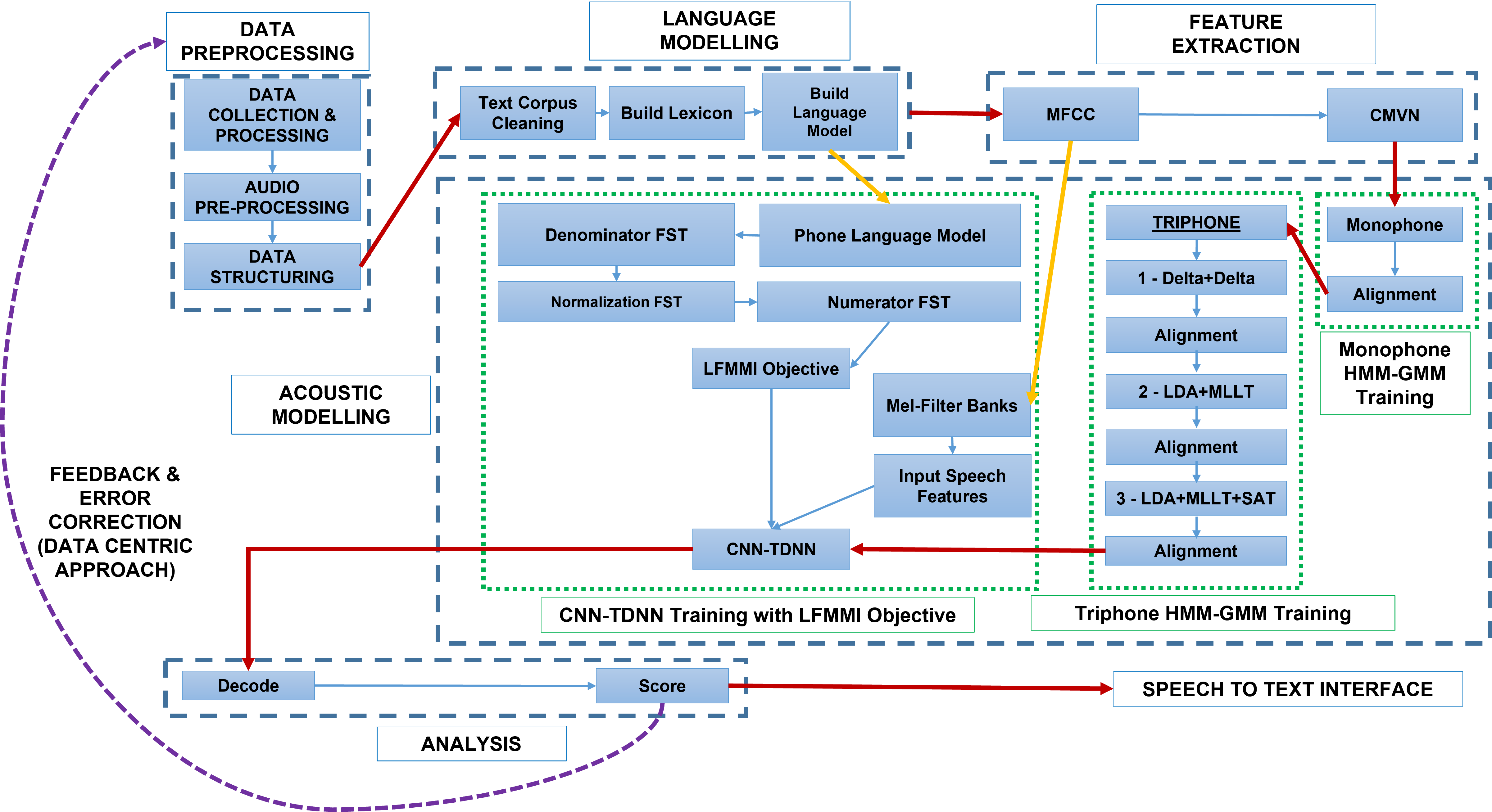}
    \caption{Workflow Summary}
    \label{fig:workflow}
\end{figure*}

For Acoustic Modeling we used a hybrid HMM-DNN in which we use HMM based state posteriors and CNN-TDNN with LF-MMI objective function which, as per our review, gives a good accuracy with less amount of data-set with noisy telephonic audio. This is followed by Analysis where Error Rates and Accuracy is computed. In case of errors or low accuracy, we go back to Pre-processing module to check for possible errors and then move on to next modules. The Final Modules is a Speech To Text Interface which is connected to the Call center to use the relevant ASR model files to give text output from input speech. The detailed summary of work flow is shown in Figure 
\ref{fig:workflow}

\subsection{Data Pre\-processing}

\begin{figure}[h]
    \centering
    \includegraphics[width=0.9\textwidth]{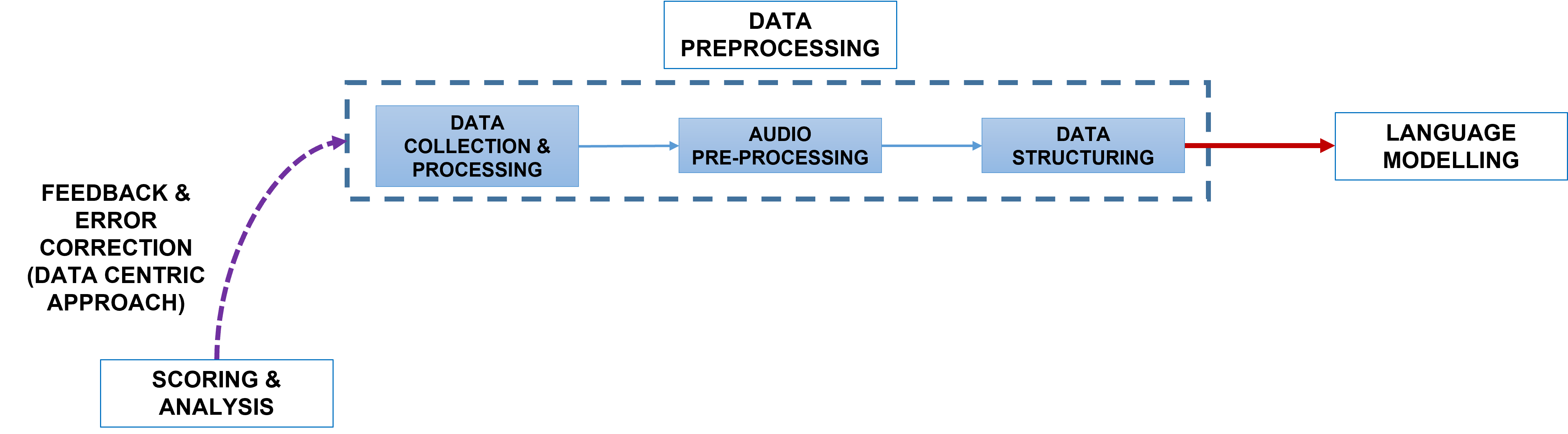}
    \caption{Data Pre\-processing}
    \label{fig:working_pipeline-1}
\end{figure}

This is the First step in training of an ASR system which comprises of three steps as shown in Figure \ref{fig:working_pipeline-1}in subsequent subsections. This process is the most time consuming but it is the most crucial one in a data-centric approach because errors in this step will propagate in the next steps which are then very hard to debug. Data is critical in AI research, and developing a strategy that prioritizes obtaining high-quality data is crucial which is why labeling quality of data primarily determines the overall quality of the machine learning model. Relevant data is not only rare and noisy, but also extremely expensive to obtain. Hence, the idea is that AI should be treated in the same way that the finest materials would be when building a house which is why evaluation at each level rather than just once. 

\subsubsection{Data Collection and Processing}
\label{sec:Data_preprocessing}


The collection and processing stage involves gathering, labelling and iterative reviewing of our data. It is important to gather as much audio data as possible at this stage weather from open source platforms or by self generated/ recorded based on given scenario. This will give the context of ASR training and is the building block for Language and Acoustic Modelling. We had various sources of data which are enlisted in Section \ref{sec:experimental-results}.

\subsubsection{Audio Pre-Processing}

The sampling rates of available audio files was largely mixed i.e. some had 8KHz and others had 16KHz frequency. We chose a higher sampling rate of 16000 Hz to have more information of speech signal is saved in the noisy or telephonic data. Generally, for ASR training, it is preferred \cite{noauthor_kaldi_nodate} that audio files with mono channel with 8kHz or 16kHz sampling frequency are used \cite{noauthor_why_nodate}. 

Instead of removing noise to maintain speech, we made sure that each utterance had a clean and a noisy counterpart which would allow the ASR to learn how the words e.g. "Salaam" sounds like in clean and noisy environment which is why we used a higher sampling rate so that more information is retained in the audio. We required mono audio in wav format for training \cite{noauthor_kaldi_nodate} \cite{noauthor_why_nodate} because unlike mp3 format, Wav formatted files are raw audio without compression applied to them which means there is no loss of information in the audio file. Most of our data was available in mono and for conversion from stereo to mono, we used our customized script utilizing \textit{SoX} and \textit{ffmpeg} \cite{noauthor_sox_nodate}. 



\subsubsection{Data Structuring}
\label{sub:datasources}


Once Data is collected and is in the right audio format, it has to be structured properly. Directory Structure and naming conventions are very important as it helps ease the coding and debugging process. We divided our data into two types; structured and unstructured. Structured data-sets had labeling and transcriptions available whereas the unstructured data-set had no labeling and transcriptions available. Our unstructured data-set was the major challenge. We had more than 500 hours of unstructured audios of Call center which meant that atleast thrice the time would be required to structure those audio files. 

Hence, we analyzed the scenario of the Call Center and found that the calls were mostly verification calls which meant a good recognition of digits and alphabets in English and Urdu was essentially required. The normal conversational words were greetings and goodbyes which were standard in general conversation.

We selected some calls and split them into word or sentence, utterance and speaker-wise, and transcribed them to be included as part of the training set. Five splits were made speaker wise with utterances for speakers being 348, 8, 7, 55 and 12 clips. For test, 38 utterances were set aside separate to the structured data to judge the results separately. These clips were taken at random, containing out-of-vocabulary words, overlapping speech, distortion and low volume at times, just like in real-life scenario. On average utterances in audio files in all cases were 1-20 seconds, covering either single words, multiple words, or up to 3 sentences.

 
The feature extraction of the input signal is limited by the memory of GPUs in use, it is best to slice all audio files not to exceed 30 seconds, as it is a reasonable input size for batching. The total audio length for training was 6hours 47minutes and testing was 3h17m. The utterances which sliced from calls were given 0.5 \- 1 second silence in start and end to ensure that there are is no abruptness in the clips, making it easier for ASR to process and align.

\subsection{Language Model}
\label{sec:our_lang_modelling}

\begin{figure}[h]
    \centering
    \includegraphics[width=0.8\textwidth]{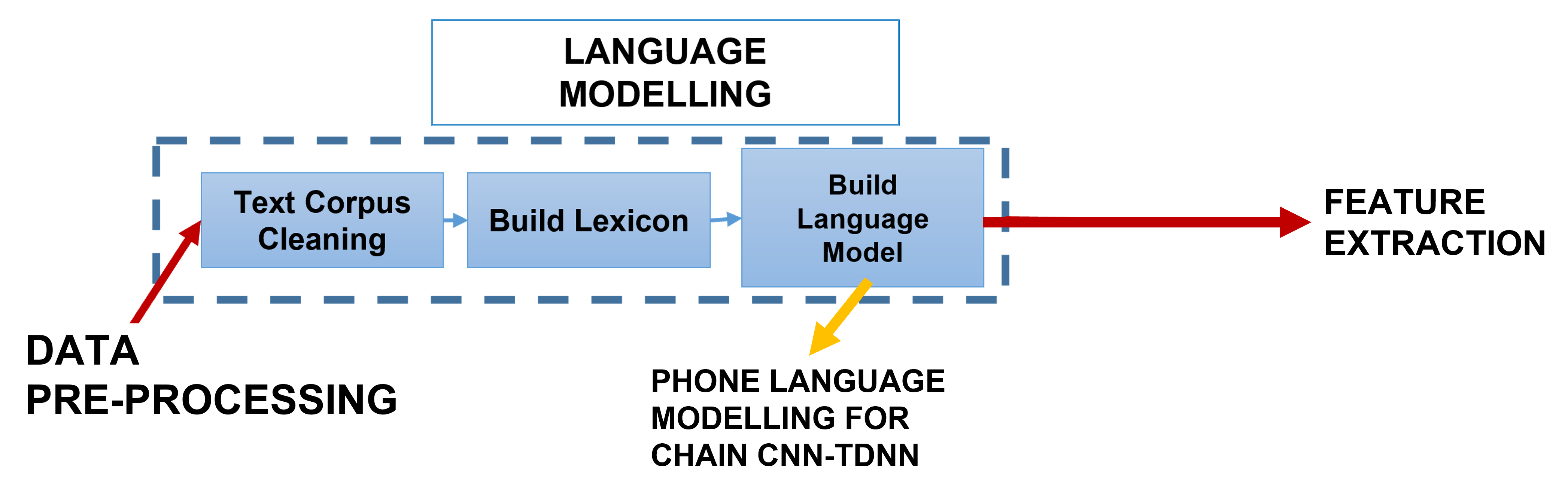}
    \caption{Language Modelling}
    \label{fig:working_pipeline-2}
\end{figure}

\begin{figure}[h]
    \centering
    \includegraphics[width=0.45\textwidth]{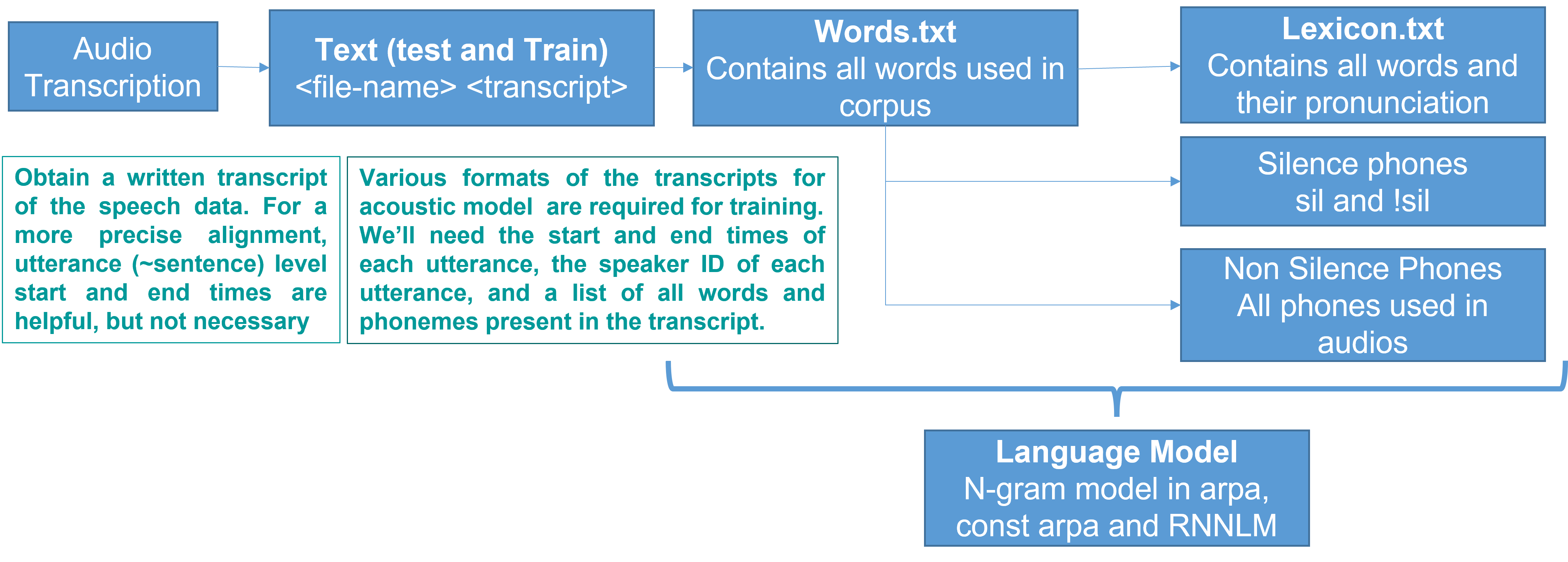}
    \caption{Language Model File Structure}
    \label{fig:lang-model1}
\end{figure}

Language Model (LM) finds the probabilities of words succeeding or preceding a specific word in a given sequence i.e. estimation of likelihood of a word-sequence $W = w_{1},...,w_{n}$ forming a valid sentence, thereby reducing the search radius of the decoders. It can be used to make decisions when the acoustic model output consists of a set of phonemes which can form various alternative sentences. Although these alternatives may be very acoustically similar, the LM selects the one that makes the most sense. LMs are usually formatted in ARPA (Advanced Research Projects Agency) format which are converted into Finite State Transducers in ASR engines. The pipline for Language modelling is shown in Figure \ref{fig:working_pipeline-2} and the preparation process is shown in Figure \ref{fig:lang-model1}. 

\subsubsection{Text Corpus Cleaning}

To prepare LM we combined the transcripts of test and train files in their respective files called \textit{"text"} which contains file-names in one column and their transcripts on the other column. We filtered the corpus and took out all unique words in the text and saved it as \textit{"words.txt"}. 

\subsubsection{Building Lexicon}
There were 7333 words, 137 phones, and 75 non-silence phones in the lexicon, each phoneme represented by a unique combination of symbols and letters. A silence phoneme, 'sil' is added to the 137 phonemes to represent the silence between hyphenated words or the beginning or end of a sentence. The mapping from speaker to utterance was done in speaker to utterance (spk2utt) and utterance to speaker (utt2spk) files.

Keeping in mind the code-switching problem we kept a single written script i.e. Roman Urdu for symbol representation. All words. whether in English or Urdu, were in Roman script which eased the lexicon building and language modelling process. The Roman Urdu script was used as symbol representations to create a dictionary that shows how words are mapped to phoneme sequences. Word to phoneme dictionary was prepared with help of \textit{lextool} by CMU \cite{cmu_cmu_nodate} and saved it as \textit{"lexicon.txt"}. Using pronunciation probabilities, multiple pronunciations of the same words were also included.  

\subsubsection{Building Language Model}

We used SRILM \cite{andreas_stolcke_srilm_2002} for language modelling and building grapheme to phoneme (g2p) sequence-to-sequence model for phonetic mappings of generated lexicon. There are various types of language modelling based on the number of phonemes considered at once. Assume we have a k-length sequence. Let $P(w_{1}, w_{2}, w_{3},..., w_{k})$ denote the probabilities assigned to the entire sequence by LM. We used a traditional n-gram model using SRILM \cite{andreas_stolcke_srilm_2002}. For an n-gram model, the probability of any word sequence $P(w_{1}, w_{2}, w_{3},..., w_{k})$ is then given as:


\begin{equation}
P(w_{1},w_{2}...,w_{k}) = \prod\nolimits_{i = 1}^{k} a_{i} P(w_{i} | w_{1},...,w_{i-1})
\end{equation}

\subsection{Feature Extraction}
\label{sec:feature-extraction}
Audio files in raw form is not understood by Computer. It has to be in certain format and some feature representation is required for the computer to make sense of the audio file and correlate it with the text transcription. Feature Extraction entails computation of features from speech wave-form containing relevant information about the linguistic content of the speech while ignoring background noise, emotions etc. The extracted features represent the phones within the words, while other signal-degrading elements such as channel characteristics and background noise are suppressed. Mel-frequency cepstral coefficients (MFCC) and Perceptual Linear Prediction Coefficients are two popular feature extraction methods (PLP). 

\begin{figure}[h]
    \centering
    \includegraphics[width=0.8\textwidth]{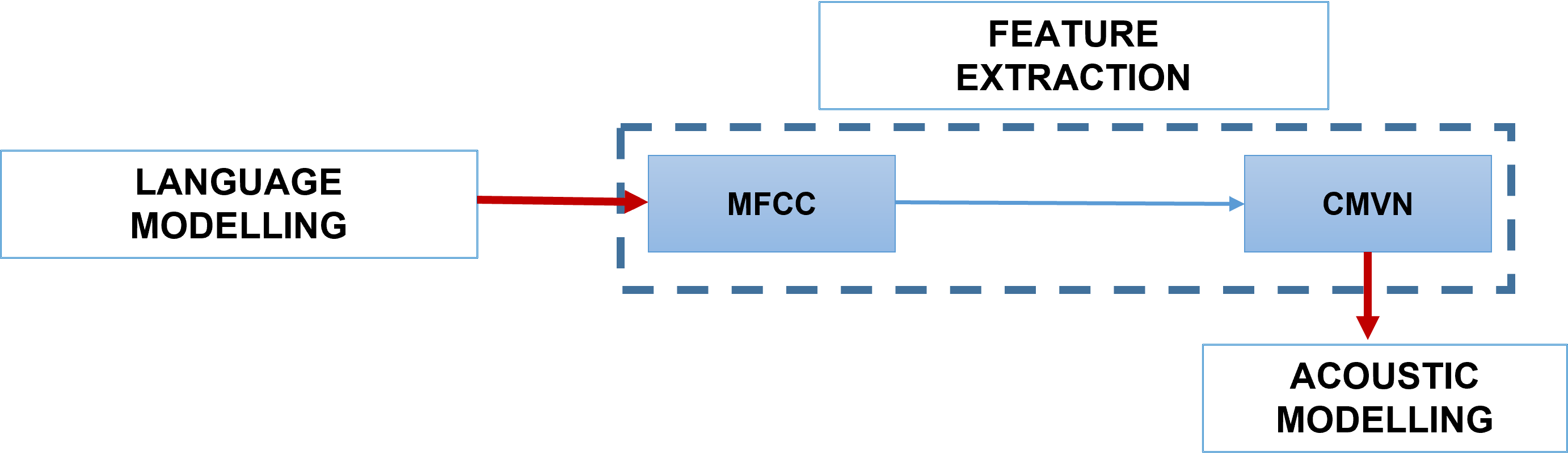}
    \caption{Feature Extraction}
    \label{fig:working_pipeline-3}
\end{figure}

The steps involved are shown in Figure \ref{fig:working_pipeline-3} and their details are in ensuing subsections.

\subsubsection{Mel-frequency Cepstral coefficients}
\label{sub:MFCC-trg}

\begin{figure}[h]
    \centering
    \includegraphics[width=0.8\textwidth]{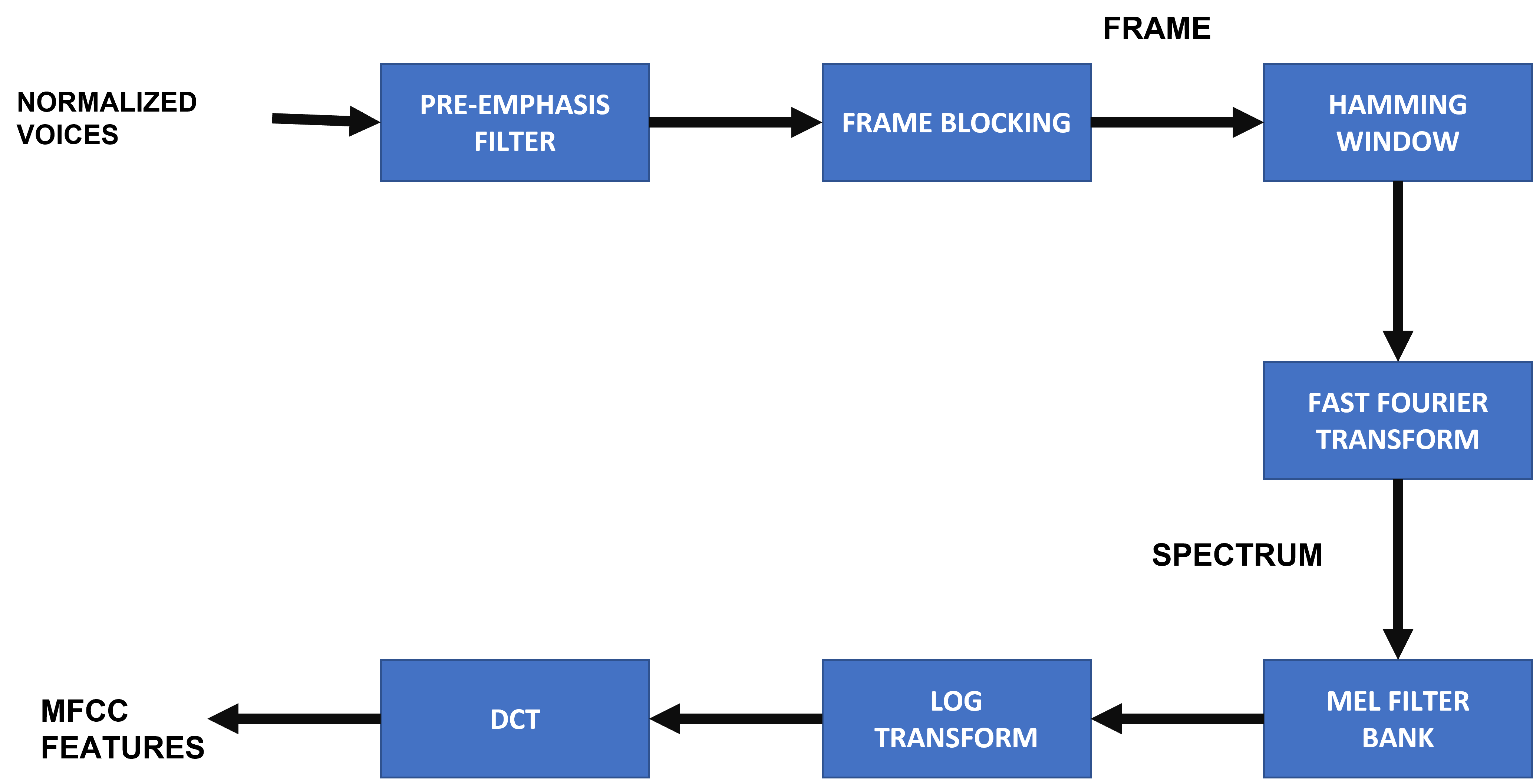}
    \caption{Computing MFCCs}
    \label{fig:MFCCs-computation}
\end{figure}

We extracted 40 MFCCs with  window size of 25ms and 10ms shift, from our training data. The MFCC extraction process begins by conversion of Analogue signals to Digital so that computers can understand them. The next step as shown in Figure \ref{fig:MFCCs-computation}, Pre-emphasis boosts the energy present in the higher frequencies to counter the spectral tilt issue which is that lower frequencies have more energy. Energy in the higher frequencies are enhanced for easy detection of third formant or F3 i.e. one of the three formants in a phoneme. 

The next step is windowing which is the process of slicing an audio waveform into small overlapping frames because each frame represents a single phoneme which is why we must carefully choose the window length. The entire waveform is divided into 20-40 ms segments. The assumption is that the signal is statistically stationary in this short segment, so features can be assumed to be constant inside this window. 

Windows overlap as they are 25 ms long and spaced 10 ms apart. In one window, approximately 400 samples are reduced to 13 Cepstral coefficients. The feature is enhanced with additional 13 delta and 13 delta-delta coefficients making a total of 39 Features. Using Hamming Window, this method smooths down the sudden edges in each frame. DFT is then applied to transfer information from the time to the frequency domain. 

After this, a set of triangular band pass filters called Mel filter are applied to the waveform after squaring the DFT output which converts information to the power spectrum. The power spectrum (or estimate of the spectral density of a signal called periodogram) is computed to recognise the frequencies present in this short segment which is accomplished through the use of discrete-time Fourier transforms.   

MFCC is still used extensively because they are easily compressible compared to Filter Banks and being de-correlated; we dump them to disk with compression to 1 byte per coefficient. It is equivalent to filter-banks times a full-rank matrix without information loss because we dump all the coefficients. It has a historical momentum which is why everyone is familiar with it. Hence there is no requirement switch to another model from a familiar one without a very strong reason. If the machine learning algorithm is susceptible to correlated input we use MFCC, a common use case of which are HMM-GMM statistical ASR models. We use Mel-scaled filter banks if the machine learning algorithm is not susceptible to highly correlated input like Neural Networks. 

\subsubsection{Cepstral Mean and Variance Normalization}
\label{sub:CMVN}
Cepstral calculation isolates the source from the filter. Pitch information is removed by MFCC features as it is not required in the Speech To Text scenario. The log spectrum contains information about the phone and the pitch. The formants that distinguish phones are identified by the peaks in the spectrum. The IFT can be applied to separate the pitch information from the formants. All that is left to do is take the first 12 independent cepstral values.


Cepstral mean and variance normalisation (CMVN) is a popular noise compensation technique and a computationally efficient noise robust speech recognition normalisation technique used in a variety of speech applications. CMVN eliminates the mismatch between training and test utterances by transforming them to zero mean and unit variance. CMVN performance degrades for short utterances due to a lack of data for parameter estimation and a loss of discriminable information because all utterances are forced to have a zero mean and unit variance hence, they are most effective for long utterances. 


\subsection{Acoustic Modelling}
\label{sec:training_the_model}

\begin{figure}[h]
    \centering
    \includegraphics[width=0.8\textwidth]{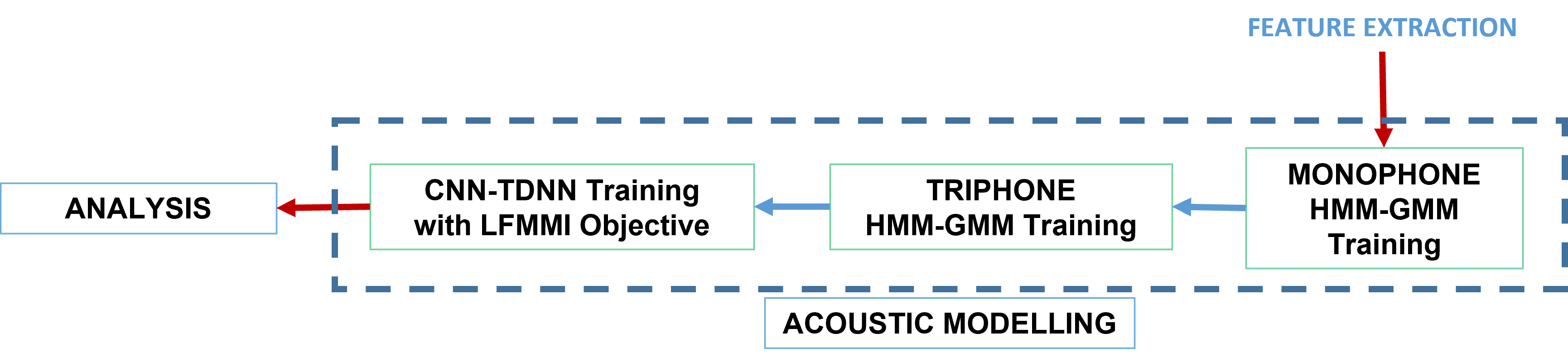}
    \caption{Acoustic Modelling}
    \label{fig:working_pipeline-4}
\end{figure}

In this section, we will discuss our Acoustic Modelling process which is shown in Figure \ref{fig:working_pipeline-4}. Acoustic models are statistical representations of the acoustic information in a phoneme which convert the audio features we created into a sequence of context-dependent phonemes and are required for both automatic speech recognition and forced alignment. 

Our dictionary contained 7333 words and 137 phonemes (75 non-silence phonemes) based on our training corpus. Alignment requires time markings on the phonemes which is why, during alignment, the system divides each audio file into equal alignments, with each division mapped to a different phoneme symbol in the sequence. Each model refines the alignments further using different training techniques before passing them on to the next stage, which is then used for recognition. The Word Error Rates are calculated Mono-phone, Tri-phone, and Chain CNN-TDNN acoustic models, to validate the ASR system. We first trained the GMM-HMM model with our training data-set before we can train the DNN-HMM model. HMM-DNN acoustic modeling consists of the steps shown in Figure \ref{fig:working_pipeline-4} are explained in ensuing subsections. 


\subsubsection{Mono-phone Training}
\label{sub:mono-phone}
\begin{figure}[h]
    \centering
    \includegraphics[width=0.8\textwidth]{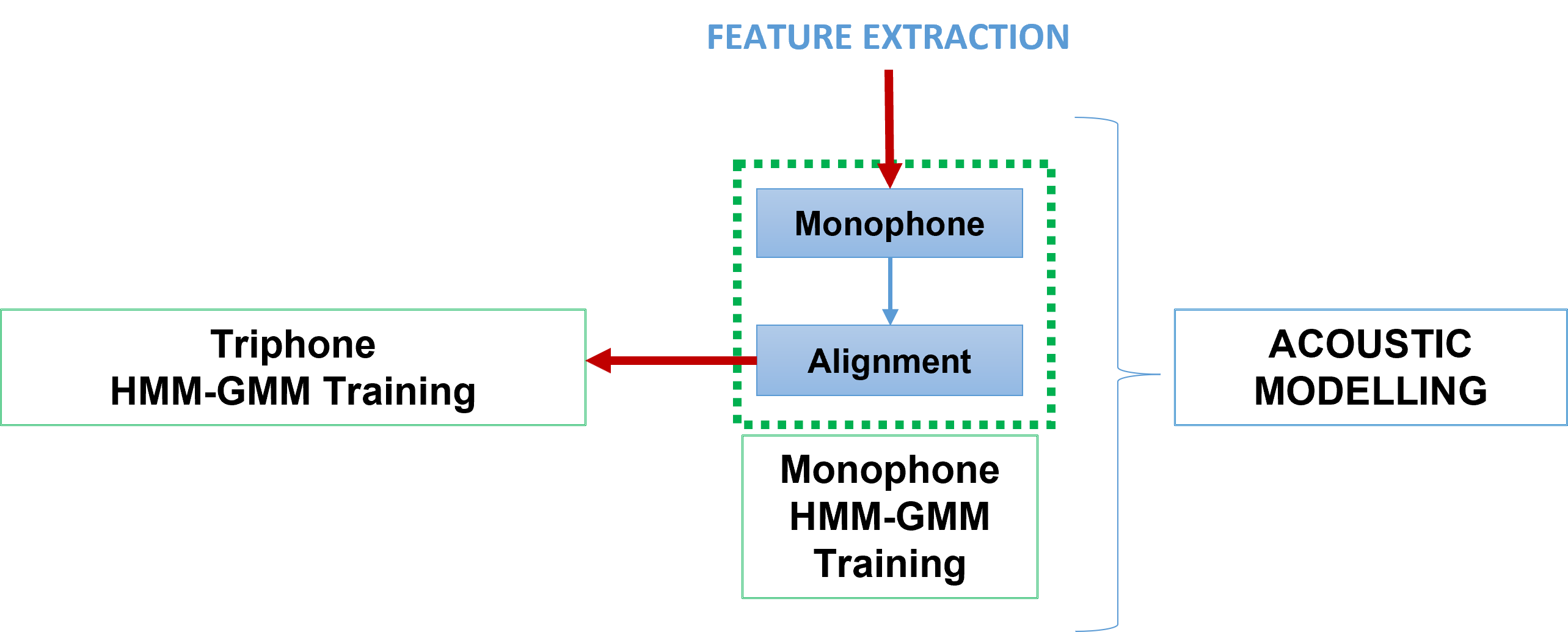}
    \caption{Monophone Training}
    \label{fig:working_pipeline-5}
\end{figure}

The first step in training Acoustic Model for HMM-DNN system is to build a simple HMM-GMM acoustic model or mono-phone model in which the HMM states model context-independent phones which is used to force-align the training data-set so that rough estimates of phone boundaries can be obtained for a more complex model. The steps are shown in Figure \ref{fig:working_pipeline-5}. 

Mono-phone model does not include any contextual information about the preceding or the following phone and is used as a building block for the tri-phone models that utilize contextual information. 

After mono-phone training, we will align the audio with the acoustic models before moving on to the next training step, tri-phone training. The parameters of the acoustic model are estimated in acoustic training steps; however, the process can be improved by cycling through training and alignment phases, also known as Viterbi training. The Expectation Maximization and Forward-Backward algorithms are related but more computationally expensive procedures.

Additional training algorithms can improve or refine the model's parameters by using the output of aligning the audio to the reference transcript with the most recent acoustic model. As a result, there will be an alignment step after each training step in which the audio and text can be realigned. 

\subsubsection{Tri-phone Training}
\label{sub:tri-phone}
\begin{figure}[h]
    \centering
    \includegraphics[width=0.6\textwidth]{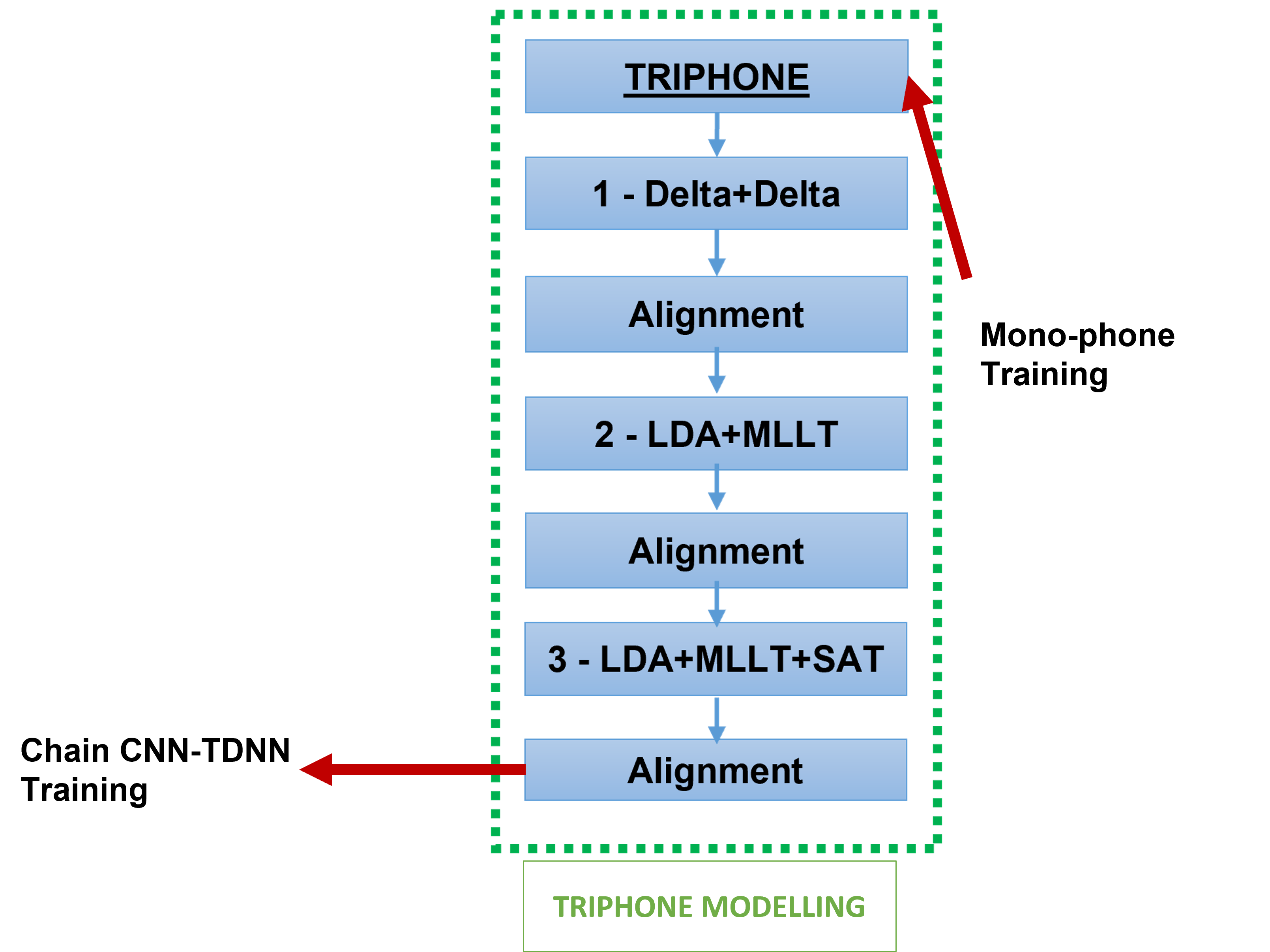}
    \caption{Triphone Training}
    \label{fig:working_pipeline-6}
\end{figure}

The next step is Triphone Training which learns the context of one phoneme before and after it, hence giving a better accuracy compared to monophone. Phonemes will vary considerably depending on their particular context, while mono phone models represent simply the acoustic parameters of a single phoneme. The tri-phone models represent a phoneme variant within the context of two other phonemes (left and right). The steps involved are shown in Figure \ref{fig:working_pipeline-6}. 

The data-set, in almost all possible cases, does not contain all tri-phone units. There are three phonemes and thus three tri-phone models, but only a subset of those will appear in the data. To gather enough statistics for the data, the unit must appear multiple times in the data. A phonetic decision tree divides these tri-phones into fewer acoustically distinct units, reducing the number of parameters and making the problem computationally feasible.

In contrast to the mono-phone model, which compares each phoneme individually and assigns weights based on the probability of match, the acoustic parameters in tri-phone models are represented for a block of three consecutive phonemes. We can assign one HMM state from a total of 49 x 3 = 147  for each tri-phone. The training data-set, however, may not contain all of the tri-phones. A decision tree is used to group the tri-phones into a smaller set of distinct units.


The tri-phone model comprises of three types training models which we trained \cite{raj_note_nodate}:

\paragraph{Delta and delta-delta training} 
Computation of delta and double-delta features, also known as 'dynamic coefficients,' in conjunction to MFCC features is done in this part. They are the signal's first and second order derivatives (features). These characteristics are then used for recognition. 

In order to recognize speech better, dynamics of the power spectrum need to be understood i.e. the trajectories of MFCCs over time for which we use delta or differential and delta-delta or acceleration coefficients. The delta coefficients are calculated using the following formula:
        \begin{equation}
        d_{t}=\frac{\sum_{n=1}^{N} n(c_{t+n}-c_{t-n})}{2 \sum_{n=1}^{N}n^2}    
        \end{equation}

The delta coefficient from frame \textit{t} is represented by $d_{t}$ and is computed in terms of $c_{t-n}$ i.e. static coefficients to $c_{t+n}$. We usually take $n = 2$. Acceleration coefficients are similarly computed using differential instead of the static coefficients.

\paragraph{LDA-MLLT} 
TDNN-based architecture use an \textit{"LDA-like transformation"} that is essentially an \textit{"affine transformation"} of the spliced input. LDA shrinks the feature space by generating HMM states for the feature vectors and the reduced feature space is used to create a transform that is unique to each speaker which is why MLLT is able to incorporate speaker independence.             

\paragraph{Speaker Adaptive Training (LDA+MLLT+SAT)} - For each speaker, noise and speaker normalisation is performed using a data transform which enables the model to compute the variance due to the phoneme rather than the background environment using its parameters. SAT also performs speaker and noise normalisation by adapting a particular data transform to each individual speaker. The data, thus, becomes more homogeneous or standardised, allowing the model to focus its parameters on estimating variance due to the phoneme rather than the speaker or recording environment. Hence, this result has the lowest word error rate. Following SAT training, the acoustic model is trained on speaker-normalized features rather than the original features. 

Audio is re-aligned with the acoustic models at the end of each tri-phone training before proceeding on to the next tri-phone model training step. In this step, the alignment algorithms include speaker-independent alignments and FMLLR. The Acoustic Model is no longer trained with the new normalised features after achieving the speaker-normalized features for SAT training. These partially speaker-independent models are used in the alignment process. Various scripts accept various types of acoustic model input but the actual alignment algorithm will always be the same. In the alignment process, speaker-independent alignment will exclude speaker-specific information. This training resulted in 13\% WER and 31.2\% SER.

\subsection{Chain Model (LFMMI Objective Function)}
\label{sec:LFMMI-chain}

\begin{figure}[h]
    \centering
    \includegraphics[width=0.9\textwidth]{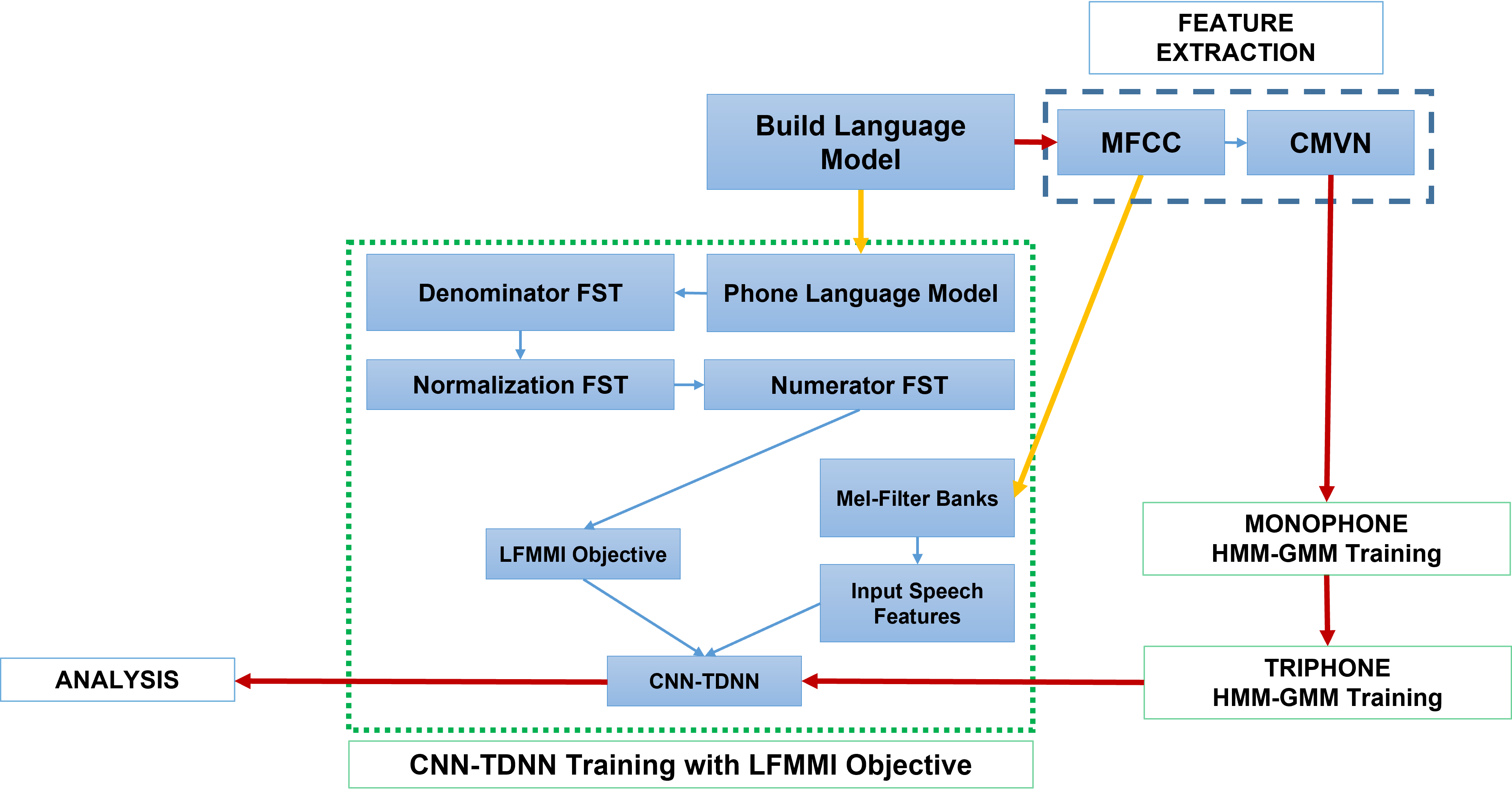}
    \caption{Chain CNN-TDNN Process flow}
    \label{fig:working_pipeline-7}
\end{figure}

The Neural Network model, which introduces the concept of unsupervised training or adaptation, eliminates the need to provide adaptation data in advance. Hence we will enhance our system by modelling the acoustic units with DNN rather than using GMMs. The neural network requires an objective function for adjustments of its parameters during training which is shown in Figure \ref{fig:working_pipeline-7}.

Discriminative objective functions like Maximum Mutual Information (MMI) or Maximum Conditional Likelihood Estimation (MCLE) objective functions are trained to maximize the gap between the correct and incorrect answers, or to differentiate between right and wrong answers instead of assigning high weights values to the correct sequences which helps training model to improve the correct output sequence prediction, while making incorrect sequences less likely. 

Deep networks excel in feature extraction and discovering correlation among them which allows exploitation of contents in making predictions which is why they can be used in ASR to classify phones based on the features extracted in acoustic frames. It is treated like a classifier using softmax to output the probability distribution $P(phone | x_{i})$. The softmax pulls up the ground truth while pulls down the others which is the same concept as Maximum Mutual Information Estimation. 

\begin{equation}
 p_{i} = \frac{e^{score_{i}}}{\sum_{c \in y} e^{score_c}}
\end{equation}




MMI objective for ASR can be expressed as \cite{noauthor_lattice_nodate}:
\begin{equation}
    F_{MMI}(\theta) = \sum_{r=1}^{R} log \frac{P_{\theta}((O_{r}|M_{Wr})P(w_{r})}{\sum_{\hat{w}}P_{\theta}(O_{r}|M_{\hat{w}})P(\hat{w})} 
\end{equation}

Where $M_{w}$ is HMM that corresponds to transcription \textit{w}. To normalise the numerator, the objective function accounts for the log-probability of complete utterance in the numerator dividing it by the log probability of all possible utterances in the denominator. Thus, the distributions with the subscript $\theta$ are trained parameterized distributions \cite{wiesner_lattice_2020}.


MMI requires First order Gradient based methods for optimization, such as Stochastic Gradient descent, which requires knowledge of the gradient of the MMI objective with respect to the parameter $\theta$. The neural network function performs forward propagation, while back-propagation computes the corresponding gradient. The state occupancies for the numerator and denominator terms must be computed for the gradient overall objective. 

Calculating the denominator sum requires summing over an exponentially large amount of word-sequences, which is impractical. Two methods can be used to approximate the sum.  
\begin{enumerate}
    \item \textit{N-best list:} This less used and crude method of approximation is computed once and used for all utterances.     
    \item \textit{Lattice structure:} It can be a word or phone based structure. A path through the lattice denotes a probable phone or word sequence. Lattices require initialization with a trained model which is a drawback, and cross-entropy trained systems are usually used for this purpose. 
\end{enumerate}

The LF-MMI objective function (or chain model) is the modified form of MMI discriminative objective function which enables ASR Training on GPU in HMM-DNN ASR Approach. It uses Neural Networks which are a different point of design in the space of acoustic models compared to Traditional ASR models. It uses a three times smaller frame rate at the Neural Network output which reduces the computational requirements significantly in test-time, making decoding in real-time  much simpler. The DNN's input features are at the original frame rate of 100 frames per second since CNNs, LSTMs, TDNNs, etc have recurrent connections or splicing inside them and are they are not purely feed-forward nets. This model uses a frame-level objective i.e. log-probability of the correct phone sequence. 

Due to the reduced frame rate, unconventional HMM topologies must be used to enable one-state HMM traversal. The chain model employs HMM's fixed transition probabilities but does not train it actually. Neural-net output probabilities can typically serve the same purpose as transition probabilities depending on the topology. 

Models are trained with a sequence-level objective function, i.e. the log-probability of the correct sequence, from the start. It is implemented in an MMI-based system without lattices on GPU by performing a complete forward-backward computation on a decoding graph derived from a phone-based n-gram language model.




\begin{figure*}[htb]
    \centering
    \includegraphics[width=0.95\textwidth]{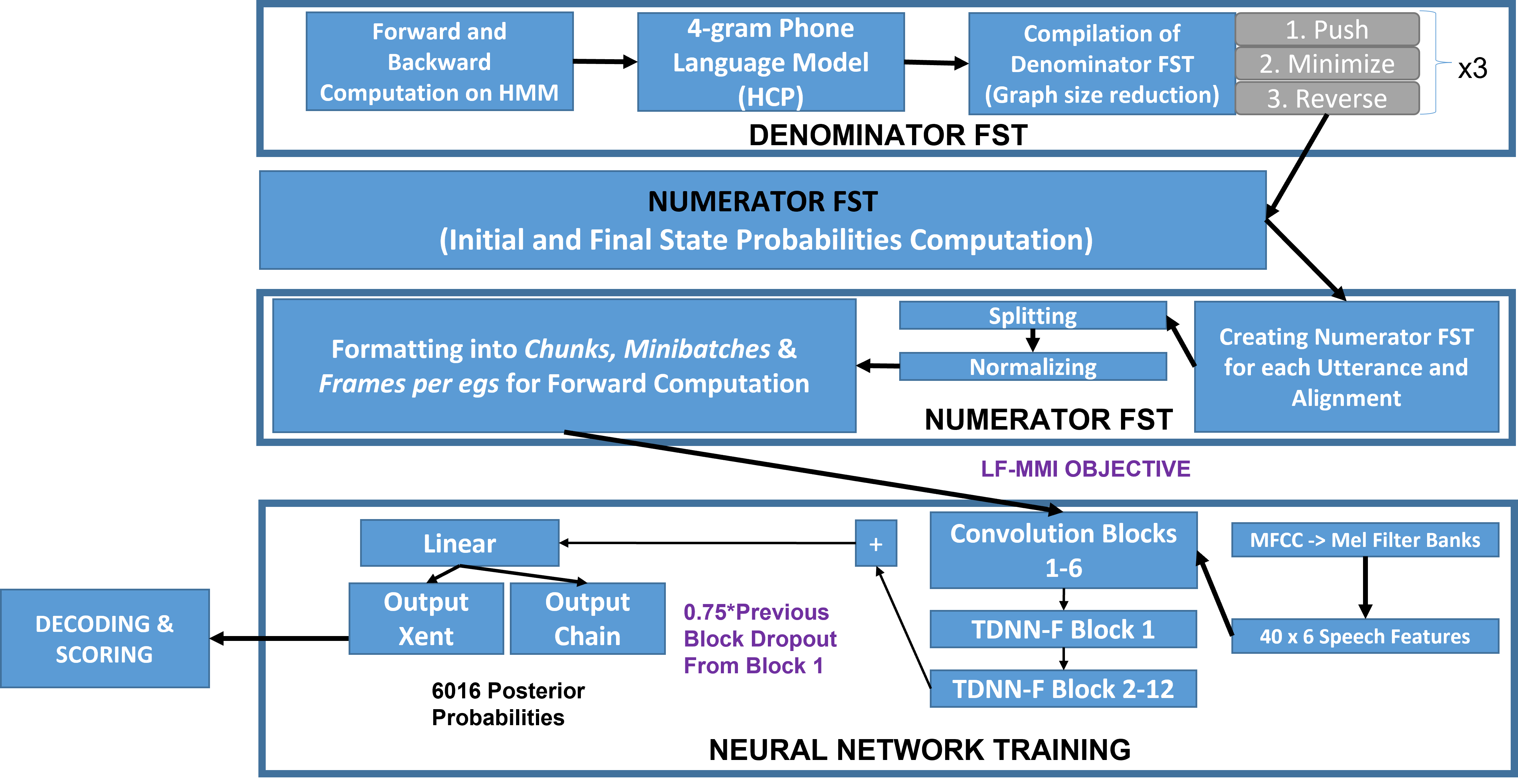}
    \caption{CNN-TDNN Training with LFMMI-Objective Function}
    \label{fig:Chain-Training-outline}
\end{figure*}

Now we will explore the key processes in the chain model as shown in Figure \ref{fig:Chain-Training-outline}.

\subsubsection{Forward and Backward Computation of Denominator FST} 

The initial probabilities of the corresponding Finite State Transducers (FSTs) requires re-adjustment since the utterances are split into 1.5 second slices. Utterances are sliced off often from the middle. This initial probability is calculated by performing the HMM for 100 time-steps beginning with the initial state and then averaging the distribution of states over these 100 time-steps. 

The HMM should be traversable in a single transition of a model at the normal frame rate instead of three transitions. The current topology contains a single-occurring state, followed by a state that can appear zero or more times. The state-clustering is obtained similarly to GMM-based models, but with a different topology for converting the alignments to the new topology and frame-rate.

The chain model training process, which is a lattice-independent version of MMI, employs a forward-backward algorithm over an HMM created from a phone-level, and the decoding graph is used to obtain denominator state posteriors. To obtain the numerator state posteriors, a similar forward-backward algorithm is used, but it is restricted to transcript-corresponding sequences. Forward and backward computations are computed in these two situation:
\begin{enumerate}
    \item During training time for computation of numerator FST and denominator FST state occupancies which are the product ($\alpha \beta$) of any states obtained using forward ($\alpha$) and backward algorithm ($\beta$) respectively.
    \item During Testing time Veterbi Decoding while pruning FST some type of forward looking is required to avoid pruning paths that can have higher probability at later stage using $\beta$ as a proxy look-ahead. 
\end{enumerate}

The gradient descent requires computation of two sets of posterior probabilities for Numerator graph specific to an utterance and Denominator graph encoding every possible word sequences which will be the same for all utterance in LF-MMI unlike the lattice-based MMI. 

A derivative of the form \textit{"numerator occupation probability - denominator occupation probability"} for each output index or pdf-id (probability density function ID) of the Neural Network is computed and propagated back into the Neural Network and the Forward-backward over an HMM for the denominator is done as part of the computation. The labels or pdf-ids are associated with the arcs rather than the states because it is represented as a finite state acceptor. Although this is not a true traditional HMM, it is still regarded as one due to the use of the forward-backward algorithm (Denominator FST) to obtain posteriors.

Like MMI, numerator and denominator \textit{'occupation probabilities'} computed and the difference between the two is used in the derivative computation. Normalization of DNN outputs is no longer required to sum to one on each frame as it makes no difference. A modified HMM topology is used because of frame rate of \textit{1 frame per 30 ms} \cite{daniel_povey_kaldi_nodate}. 

The forward and backward computations for numerator FST are on CPU since they are small and for denominator FST, these computations are performed on GPU but not on log domain since computation of log several times slows the process. The objective function values can occasionally become bad for which the objective function is kept within range of [-30,30] \cite{raj_experiments_nodate}. To reduce the maximum memory usage, the denominator computation is performed before the numerator. This algorithm is also prone to numeric overflow and underflow, which can be avoided by multiplying the emission probability of the frame by the normalisation factor $\frac{1}{alpha(t)}$.

\subsubsection{Phone LM \& Denominator FST} 
If the denominator is represented as a graph and it is fitted in the GPU allowing computations to be efficiently performed for which some modifications are done.

A Phone LM instead of Word-based LM is built to reduce size of graph significantly because there are far fewer possible phones than there are words. Phone LM is derived from phone alignments in training data making it an un-smoothed language model, which means there is no reverting to lower order n-grams.

Transition to some LM states that are removed go to lower order n-gram states. Smoothing is avoided to reduce number of arcs in compiled graph after phonetic context expansion. There is a requirement for maintenance of 2-phone history which is why a 4-gram language model is used for chain training and we never prune LM states below the trigram level which not pruned because any sparsity in which trigrams are permitted tends to reduce the size of the compiled graph.

The standard 3-state left-to-right HMM topology, commonly used in ASR, cannot be used because the entire HMM is required to be traversed in a single frame because DNN outputs are computed at a third of the standard frame-rate, which is accomplished by decreasing the frame-shift to 30ms from traditional 10ms. Training such a system with the MMI objective necessitates accurate computation of the objective and its derivative. 
   
    
    

\subsubsection{Compilation of the Denominator FST} 
Phone LM is expanded into an FST with pdf-ids as the arcs, through a process akin to a normal decoding-graph compilation, but without using a Lexicon, at end of which the transition-ids are converted to pdf-ids. 

The difference lies in graph size minimization because the standard recipe calls for determination and minimization. This procedure it's variants with disambiguation symbols, do not allow graph size reduction which is why the graph-minimization procedure is three repetitions of:  
        \begin{enumerate}[label=(\alph*)]
            \item Push
            \item Minimize
            \item Reverse (The terms 'push' and 'reverse' refer to reversing the directions of arcs and swapping initial and final states.)
        \end{enumerate}

\subsubsection{Initial \& Final Probabilities} 
A starting state and final probabilities for every state are produced by the graph-creation process naturally. These are not the states used in the forward-backward procedure because these probabilities apply to utterance boundaries but it is trained on fixed-length chunks of utterance which is usually 1.5 seconds. Constraining the HMM to the initial and final states at these arbitrarily chosen cut points is not appropriate. For each state instead, we use: 
        \begin{enumerate}[label=(\alph*)]
            \item Derivation of Initial Probabilities from fixed iterations of HMM-running 
            \item Averaging the probabilities
            \item Final probabilities = 1.0 for each state 
        \end{enumerate}
                    
\subsubsection{Normalization FST} 
The initial and final probabilities are applied to the initial and final frame as part of the computation in the denominator forward-backward process. Normalisation FST is a version of the denominator FST with these initial and final probabilities. Epsilon ($\epsilon$) arcs are used to simulate initial probabilities since they are unsupported by FSTs. Probabilities will then be added to normalization FST using Normalization FSTs.   
    
\subsubsection{Numerator FSTs} 
A numerator FST for each utterance is generated to prepare the training process, which encodes the supervision transcript along with the transcript alignment to force similarity to reference alignment, allowing a little deviation, acquired from a base-line system. 

A phone is allowed  in the lattice alignment to occur 0.05 seconds prior or after its start and end positions, respectively. Alignment details must be included as training is done on fixed-length utterance segments rather than complete utterances, i.e. If we know where the transcript aligns, we can split the utterance into pieces, which is important for GPU-based training

Rather than enforcing a specific pronunciation of the training data, a lattice of alternative pronunciations of the training data is used, which is generated by a lattice-generating decoding procedure that uses an utterance-specific graph as our reference decoding graph, generating all pronunciation alignments within a beam of the highest-scoring pronunciation.

\subsubsection{Splitting the numerator FST} 
Fixed sized utterances of 1.5 seconds length are used, necessitating division of the numerator FSTs into fixed-size segments. This is simple because the numerator FSTs remembers and encode time-alignment information and  have a structure that allows association of any FST state with a specific frame index. At this time, there are no costs in the numerator FST. It is only viewed as encoding a path constraint, so there is no need to decide how to split up the costs on the paths.

\subsubsection{Normalizing the numerator FSTs} 
Normalisation FST is merged with split-up pieces of numerator FST to ensure that the denominator FST costs reflect in the numerator FST which in turn ensures that objective functions are never positive and making their interpretation easier. 

This prevents the objective function from ever decreasing by preventing the numerator FST from having state sequences that the denominator FST would find unacceptable while also protecting against the possibility of numerator FST containing state sequences not accepted by denominator FST, which allows the objective function to indefinitely increase. 

The Lattices may contain n-gram phone sequences which were not seen in training since Phone LM has no smoothing and is estimated from 1-best alignments. Normalization process infrequently generates empty FSTs which can happen in following scenarios: 
        \begin{enumerate}[label=(\alph*)]
            \item The lattice consists of tri-phones absent in the 1-best alignment that was used to train the Phone LM and, 
            \item No alternative paths are available at that point in the lattice to compensate for the resultant failed paths which is a possibility because the lattice-producing alignment and 1-best alignment selected different pronunciations of the same word which is why These utterance fragments are simply discarded.
        \end{enumerate}

The initial probabilities are now have different values which becomes a problem for chunk-level FSTs. To approximate this, the HMM running is done for a few iterations and then the probabilities are averaged to be used as the initial probability of any state. Thus Normalization FST seems to work despite being a crude type of approximation. The numerator FST is composed with the normalization FST to ensure that objective function value is always negative for easier interpretation and that the numerator FST is free of sequences not permitted by the normalisation or denominator FST, which commonly happens because the sum of the overall path weights for such sequences is dominated by the normalization FST part. 
        
\subsubsection{Formatting numerator FSTs} 
Pdf-ids can not be directly used as weighted acceptors in the numerator FSTs because they could be zero which is treated differently by Open-Fst i.e. as epsilon. Hence, numerator FSTs' weighted acceptors correspond to \textit{"pdf-ids + 1"}. Instead of storing an array of separate numerator FSTs, they are appended together to form a longer FST when mini-batches are formed. This enables a single forward-backward algorithm to be applied to every utterance in the mini-batch, directly calculating the total numerator log-probability. 

\subsubsection{Chunks-Size} 
The number of output frames for each chunk of data that we analyse and evaluate during training or decoding is known as the chunk-size. There is a mechanism for variable chunk sizes that permits reasonably large chunks while preventing data loss from files that are not exact multiples of the chunk size. The first specified chunk size is called primary chunk-size which is unique because we can choose no more than two non-primary chunk sizes for any given utterance and the remaining chunks must be of the primary chunk size. Determining the optimal split of a file of a given length into chunks is made easier by this restriction, allowing the biasing of chunk-generation to chunks of a specific length. The number of frames per chunk influences the speed or latency trade-off but not the results.

\subsubsection{Mini-Batches} 
A larger mini-batch size is generally more efficient because it interacts well with optimizations used in matrix multiplication code, particularly on GPUs, but it can lead to update instability if it is too large and the learning rate is too high. 


To reduce the size of the denominator FST for training on the GPU training is done on 1 to 1.5 seconds chunks, instead of the complete utterance. However, the transcript needs to be broken up to do this and 1-second chunks may not coincide with word boundaries. Utterances are divided into given fixed-length chunks of speech to train on mini-batches of length 1.0 seconds. Shorter utterances are removed, while longer utterances are divided into chunks with small gaps or overlaps between them.
        
The acoustic models usually require left or right frames for acoustic context which is added but the context is added after the chunks have been determined. Usually the mini-batch size is a power of two, and is possibly limited by GPU memory constraints. We used 64 mini-batches based on our data-set. The alpha probabilities in forward-backward computation consume the most GPU memory. After the 3-fold sub-sampling, we have 50 time steps with a 1 second chunk.

\subsubsection{Frames per Egs} 
The entire data-set is randomised at the frame level for the most basic types of networks, such as feed-forward networks or TDNNs trained with the cross-entropy objective function, and training is done on one frame at a time. 



\subsubsection{Avoiding Over-fitting}
LF-MMI uses following techniques to avoid over-fitting: 
\begin{enumerate}[label=(\alph*)]
    \item \textit{L-2 regularization on neural network outputs:} This forces the weight parameters to decay or approach zero but become exactly zero. It is also called Ride Regression or Weight Decay. Smaller weight parameters makes neurons negligent which reduces neural network complexity and hence, over-fitting. 
    \item \textit{Separate classifier Output Layers:} LFMMI uses two separate output layers for classifier, one trained with Maximum mutual information estimation (MMIE) and the other with cross-entropy so that the overlapped layers will be trained with both objectives to reduce over-fitting that is caused by a single objective.
    \item \textit{Multitask training with the cross-entropy objective:} The forward probabilities at each time step in the numerator graph are used as soft targets instead of the usual hard targets.
    \item \textit{Leaky HMM:} This permits small transition probability between any two states is allowed allowing for gradual context forgetting. 
\end{enumerate}

\subsection{Neural Network Training with CNN-TDNN}

Once we have the Lattice Free Maximum Mutual Information (LF-MMI) objective function computed, the next process is Chain CNN-TDNN training which is shown in Figure \ref{fig:working_pipeline-8}. For The Neural Network training part, the 1500ms speech segments with 40 Mel filter-banks features and 200 iVectors are the input and structure of the network is CNN-TDNN as outlined in Figure \ref{fig:Chain-Training-outline}. The LF-MMI here is the training objective function which defines how the weights are set during the training. 



Time-Delay Neural Network (TDNN), captures long term temporal correlations between speech frames i.e they are time-dilated 1-Dimensional CNNs. It is a time-domain convolutional network that models temporal dependencies which is easier to parallelize in contrast to a recurrent network and is comparable to feed-forward DNN in terms of training time \cite{noauthor_tdnn_nodate}. It has special characteristics of shift in-variance just like CNN. TDNN can learn short intervals contexts between input features like MFCC and of longer intervals in upper layers i.e. lower layers learn short input contexts, and higher layers learn long input contexts \cite{liu_time_2019}.

CNN-TDNN is a variant of simple TDNN model and it is part of a multi-component system comprising of a hybrid HMM-TDNN based acoustic model, a phonetic model and an n-gram language model. Optionally for re-scoring, a complex n-gram or NN-based LM can be used. In Chain or LFMMI model with CNN-TDNN, 4-gram model is used. 

Mel-filter-banks are used in this implementation instead of MFCCs, in terms of network input features. Here the final features are organized as a matrix unlike simple TDNN where the input features are represented as a vector. 

The input in the CNN-TDNN network comprises of two feature types organized in a 40x6 matrix of speech features \cite{georgescu_performance_2021} which includes 40-dimensional Mel-filter-banks extracted from 25 ms length frames and 200-dimensional and 10 ms i-vectors-shift calculated from chunks of consecutive 150 frames.

TDNNs are one dimensional or temporal CNNs whereas CNN works on time as well as frequency dimension. Note that unlike TDNN, CNN-TDNN requires Mel filterbanks instead of MFCCs because the CNN layer needs to have a meaningful non-time dimension to operate on which is why we want our signals to be in  frequency space or dimension. They are still dumped to disk as MFCCs during the NN extraction and training process but the DCT part of MFCC is inverted or reversed to get Filterbanks. 

The neural network component acoustic model of CNN-TDNN resembles TDNN, the key difference being placing of few CNN layers prior to the time-delay layers, acting as a front-end block. Convolutional layers have the property of annihilating small variations in the spectral domain due to their structure provided by local connectivity, weight sharing, and pooling. These variations are induced by both the speaker and the acoustic environment in which the speech takes place. 

The input for Conv. Block 1 are three matrices of speech features i.e. the features for the current, previous, and next acoustic frames, or a feature volume of 6 x 40 x 3 equivalently. Time and feature space convolutions are performed using 64 filters of size 3 x 3, and the output is 64 x 40 x 1 volume. The Second Convolutional Block's input comprises of three time consecutive volumes as the first Convolutional Block's output, that are spliced together to form the 64 x 40 x 3 feature volume. The second Convolutional block performs time and feature space convolutions with another 64 3x3 filters and gives 64 x 40 x 1 volume as output. More filters are used in Convolutional Blocks 3–6, ranging from 128 to 256, while feature volume size is kept constant by reducing the height from 40 to 20, and then to 10, finally generating a 2560-dimensional output that is passed on to the next time-delay blocks from whereon the processing is same as TDNN.

Adding CNNs before TDNN can possibly help improve performance in noisy environment. TDNNs are basically time-dilated 1 Dimensional CNNs and if the input features are noisy, having a few CNN layers (2D CNN) at the input would help the model learn some non-linear transformations which can reduce the effect of the noise on speech input. 

Front End CNN Blocks are followed by 12-blocks of factored TDNN i.e. TDNN-F. TDNN-F Block 1 processes the current time frame only, while the rest perform temporal convolution over $t-3$, $t$, and $t+3$  time indexes. The linear layer is formed by splicing the input vectors at time indexes $t-3$ and $t$ are spliced together to form the linear layer. The affine layer is formed by splicing the input vectors at time indexes $t$ and $t+3$.  

TDNN employs a sub-sampling technique to compute hidden activations only at specific time steps which avoids redundancy because the large overlap of contexts leads to highly correlated neighbouring activations. A larger context on the left side is usually found to be optimal for online decoding. This way, Model size and training time are reduced and are therefore faster than LSTM. TDNNs can learn local correlations between speech frames. Bottleneck layers, which are initialized by Singular value decomposition (SVD), constrains rank of weight matrices which help in regularization and reduces number of Multiplications $M \times N -> (M+N) \times R$. 

TDNN structure comprises of additional helps reduce spectral and temporal variability. The TDNN network's sub-sampling mechanism is similar to a convolutional operation that allows gaps in the convolutional filter which is another network variant based on the purely TDNN approach, with a couple of stacked convolutional layers added before with a primary function to further process acoustic features and act as a feature processing front-end. The output blocks use cross-entropy and chain loss functions in TDNN and CNN-TDNN both. The CNN-TDNN and simple TDNN architectures have identical Neural Network output blocks, which are represented by 6016-dimensional posterior probabilities of acoustic states, while the language model of 200K words provides the system output. 

We used 10 Epochs with 10 hidden layers while reducing the learning rate geometrically from initial learning rate of 0.00015 to the final learning rate of 0.00015, and then keep it fixed at –final-learning rate, for extra number of 5 epochs. Maximum parameter change value was 2.0 along with Initial Number of jobs =3 and Final Number of jobs= 16 for trainer optimization. We provided the forced alignments from HMM-GMM Tri-phone to our LF-MMI or chain CNN-TDNN pipeline. The training resulted in 5.2\% WER and 20.45\% SER.

\subsection{Decoding}
\label{sub:decoding}

Decoder algorithms efficiently search the hypothesis space by combining AM and LM predictions and output the most likely text-string for a given speech input file. Decoder aims to combine these various models, given the input sentence, to estimate the likelihood of a sound sequence already observed in the speech database. The system then searches through all sentences in the space, selecting the sentence with the highest probability of being the input source sentence. The solution for the \textit{search} or \textit{decoding problem} for a massive set of all English sentences requires an efficient algorithm which only searches through sentences which are more likely to match the input instead of searching through all possible English sentences.

\begin{figure}[h]
    \centering
    \includegraphics[width=0.6\textwidth]{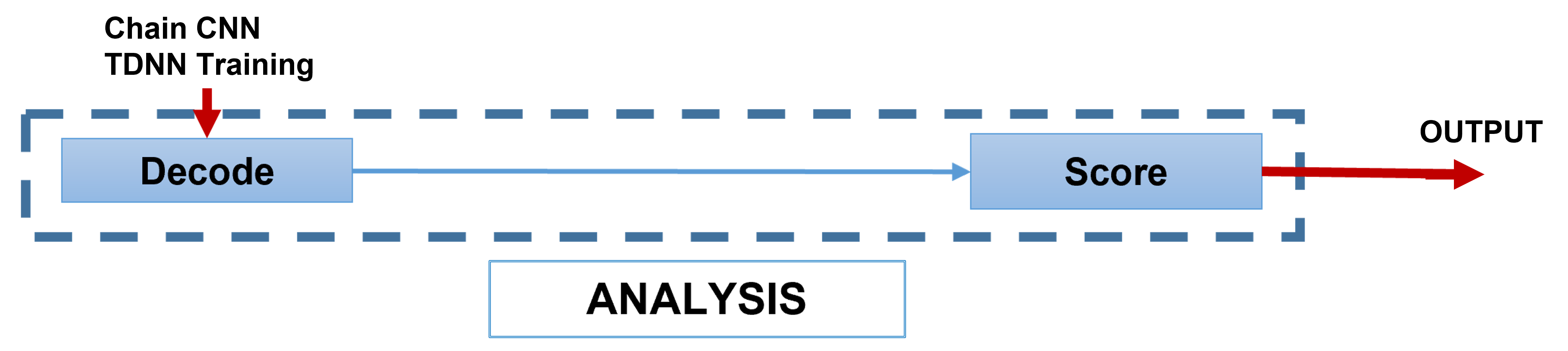}
    \caption{Analysis Process flow}
    \label{fig:working_pipeline-8}
\end{figure}

After training, we need to analyze the model by testing it with unseen data during training. For that we decode the graph files which are produced at end of each training i.e. mono-phone, tri-phone and CNN-TDNN. This is then scored and evaluated. The steps are explained in subsequent subsection and shown in Figure \ref{fig:working_pipeline-8}

\subsubsection{Mono and Triphone (HMM-GMM) Decoding}
Weighted Finite State Transducers are used in the HMM-GMM decoding algorithm, which provides graph operations for acoustic modelling. These decoding graphs are assigned numerical values known as pdf-ids that represent context-dependent states. 

Pdf-ids are numerical values of context dependent states given by decoding graphs formed by the decoding algorithms which uses WFSTs to provide graph operations used in acoustic modelling. WFST are used in HMM-GMM models and can also be used along with Deep Neural Network Classifiers.

Since different phones can have the same pdf-ids, Transition-ids are used to encode phone member pdf-ids and use arc (transition) within the topology specified for that phone. Thus, decoding is carried out on these decoding graphs HCLG, which are built from simple FST graphs. 

\begin{center}
HCLG = H ◦ C ◦ L ◦ G    
\end{center}

The ◦ symbol denotes associative binary operation of composition on FST. \textit{H} for HMM state graph or definitions, \textit{L} for Language Models or Lexicon, \textit{G} is the acceptor which is used to encode the grammar and \textit{C}, the context dependency are combined to form HCLG (saved as HCLG.fst). 
\vspace{11pt}

\subsubsection{Chain Model Decoding}

The traditional "H ◦ C ◦ L ◦ G" is a composed transducer used to decode audio. This WFST can be integrated with the deep network classifier. It is a big complex network which can be trained like DNN or DL using back-propagation without the requirement of introduction of a lattice for denominator approximation. 

The numerator and denominator state sequences are encoded as Finite State Transducer corresponding to the HMMs. The objective function overall is the difference of these FSTs in log-space which is built as the HCLG FST decoder. 

The graph becomes too big to fit into GPU memory if the FST comprises of all possible word-sequences which is why phone-level LM is used which contains fewer possible sequences. L-graph is not needed since we are using Phones instead of words. Phone-LM \textit{"P"} is a 4-gram model with no back-off less than 3-gram to ensure that tri-phones not seen in training are not generated and is constructed to minimize the graph size. Removing low-count 4-gram states completely will limit the number of states. 

Thus final graph is \textit{HCP} (where P = phone LM) file instead of \textit{HCLG}. Once HCP graph is composed, a different kind of \textit{graph minimization technique} is done by three iterations of the operations of Pushing the weights, graph minimization and reversing the arcs and swap initial and final states.

The numerator FST encodes utterance-specific alternate pronunciations of original utterance transcript. This lattice is converted into an FST with a 0.05-second error window from the phones' position in the lattice, limiting the time at which the phones can appear which is followed by conversion into an FST with labels that are pdf-ids or neural network outputs. This FST is used to extract fixed size chunks for training chunks in the denominator FST. 

The numerator FST contains the lattice for only one utterance, broken into chunks of fixed length, making it easily to calculate. It has H, C, and L without \textit{"G"} because the utterance is known. 

Phones can be taken at the numerator lattice's output and everything can be projected on the input to compose the numerator lattice with HCP. Since the numerator lattice changes with each utterance, this composition must be performed for each utterance. However, almost all paths in HCP are removed, and the final FST is very small because the numerator lattice is small and composition is analogous to intersection.

\subsection{Scoring}

Scoring is done from decoding of the graphs that are generated as a result of HMM-GMM and Neural Network Training. There are some evaluation methods for ASR which we will explore first followed by scoring process.

There is no single fixed method of evaluating ASRs, although Word Error Rate is the most popular mainstream metric for ASR Evaluation \cite{maglogiannis__2020}. We will examine some popular evaluation metrics before discussing our selection of ASR Evaluation metrics for our particular scenario.

\subsubsection{Word Error Rate} 
WER is a popular ASR comparison index and is used to rate ASR performance accuracy by comparison of hypothesised and reference transcriptions. It expresses the distance between the reference series and ASR-producing word sequence.   

The two-word sequences are first aligned using a string alignment algorithm based on dynamic programming. After alignment, total substitutions (S), insertions (I) and deletions (D) are determined. The WER is calculated by dividing the number of errors (defined as substitutions, deletion and insertion) by the number of words (N) in the reference. 

Substitution occurs when a word in the reference sequence is transcribed as a different word. When a word is completely absent from the automatic transcription, it is labelled as deleted. Insertion is the appearance of a word in the transcription that has no counterpart in the reference word sequence. WER is defined \cite{morris_wer_2004} as:
    \begin{equation}
         W.E.R = \frac{S+N+I+D}{N_{1}} =\frac{S+I+D}{H+S+D}   
    \end{equation}

Where Total Entries = I, Total Deletions = D, Total Substitutions = S, Total Successes = H and Total Reference words = $N_{1}$.

WER has some drawbacks despite its popularity. It is not a true percentage because there is no upper limit. When S = D = 0 and we have two insertions for each input word, then I = $N_{1}$ (that is when the length of the results exceeds the number of words in the prompt), i.e. WER equals 200 percent. Thus, it only takes in comparison to other systems how good ASR performs. WER may exceed 100 percent in noisy scenario because insertions are given far more weight than deletions.

\subsubsection{Character or Phone Error Rate } 
Character Error Rate (CER) or Phone Error Rate (PER) is also used by some systems which evaluate how accurately the ASR system was able to understand the word at the phones/phoneme level e.g. differentiating "there" vs "their. The character error rate is also used to check the model's accuracy. Where S = Substitutions, D = Deletions, I = Insertions and N = Number of Letters in a Single Word, the formula for calculating the character error rate is as follows.:
        \begin{equation}
             C.E.R = (S + D + I) = N     
        \end{equation}
        
\subsubsection{Sentence Error Rate } 
SER is another metric that is sometimes used to assess the ASR performance. The percentage of sentences with at least one error can be calculated by SER.

\subsubsection{Word Information Lost } 
WIL is a simple approximation to the proportion of word information lost that avoids the issues associated with the RIL measure that was proposed many years prior.

\subsubsection{Relative Information Lost } 
RIL metric was not widely adopted because it is not as straightforward as WER and measures zero error for any one-to-one mapping between input and output words, which is not acceptable.

\subsubsection{Match Error Rate } 
MER is the proportion of I/O word matches (errors) i.e. the likelihood of a given match being incorrect. MER's ranking behaviour falls somewhere between WER and WIL \cite{morris_wer_2004}.
        \begin{equation}
            M.E.R = \frac{S+D+I}{N=H+S+D+I} =1-\frac{H}{N}   
        \end{equation}
        
\subsubsection{Position-Independent Error Rate } 
ASR Evaluation also correlates with the automatic metric PER (Position-Independent Word Error Rate). When the words in the two sentences (hypothesis and reference) are compared, PER is always less than or equal to WER. The set of words in a hypothesis sentence that do not appear in the reference sentence is referred to as the hypothesis per (Hper). The set of words in a reference sentence that do not appear in the hypothesis sentence is denoted by reference PER (Rper), which is similar to recall. Hper and Rper primarily aim to identify all words in the hypothesis that do not have a counterpart in the reference and vice versa \cite{maglogiannis__2020}.

\subsubsection{Calculation of Scores }

We chose WER and SER for evaluation of our ASR. We had an option of using CER is also there but it is irrelevant for us since we are more concerned with the accuracy of words and sentences. The Urdu word \textit{"Janbaaz"} can be written as \textit{"Janbaaz"} or \textit{"Jaanbaz"} and will be spelled pretty much in the same way. As long as the overall word recognition is accurate and the sentences are accurately transcribed, our ASR is good for our requirement which is why we used WER and SER for our scoring.

The HMM-GMM part of the training produces HCLG file from which scoring is done based on chosen evaluation metric (as shown in previous subsection) by calculating insertions, deletions and substitutions from reference scripts. In the LF-MMI model, neural networks are a box that scores the gradient function's emission probability that we train. Every HMM state corresponds to pdf-id, and the neural network must provide a score for each pdf-id for each frame in the output. 

Hence input frames chunk having width \textit{"w"}, the \textit{"N x w"} dimension matrix is the Neural Network output, where \textit{"N"} corresponds to total \textit{pdf-ids}. This matrix can be represented as an FST also called as scoring FST or sausage lattice, with nodes representing frames, arcs representing \textit{pdf-ids}, and weights on the arcs representing neural network scores. The Scoring FST is composed with HCP to get final total score.

The Scoring FST provides the acoustic score, and the graph provides the graph score. The neural network is trained with lattice posteriors generated by forced alignment using a trained GMM-HMM model. The best HMM-GMM acoustic model (Triphone-3) is used to force-align the training data in order to produce high-quality alignments for the Neural Network model which in our case in CNN-TDNN. 

\subsection{Speech To Text Interface}
\label{sec:STT-interface}

\begin{figure}[h]
    \centering
    \includegraphics[width=0.9\textwidth]{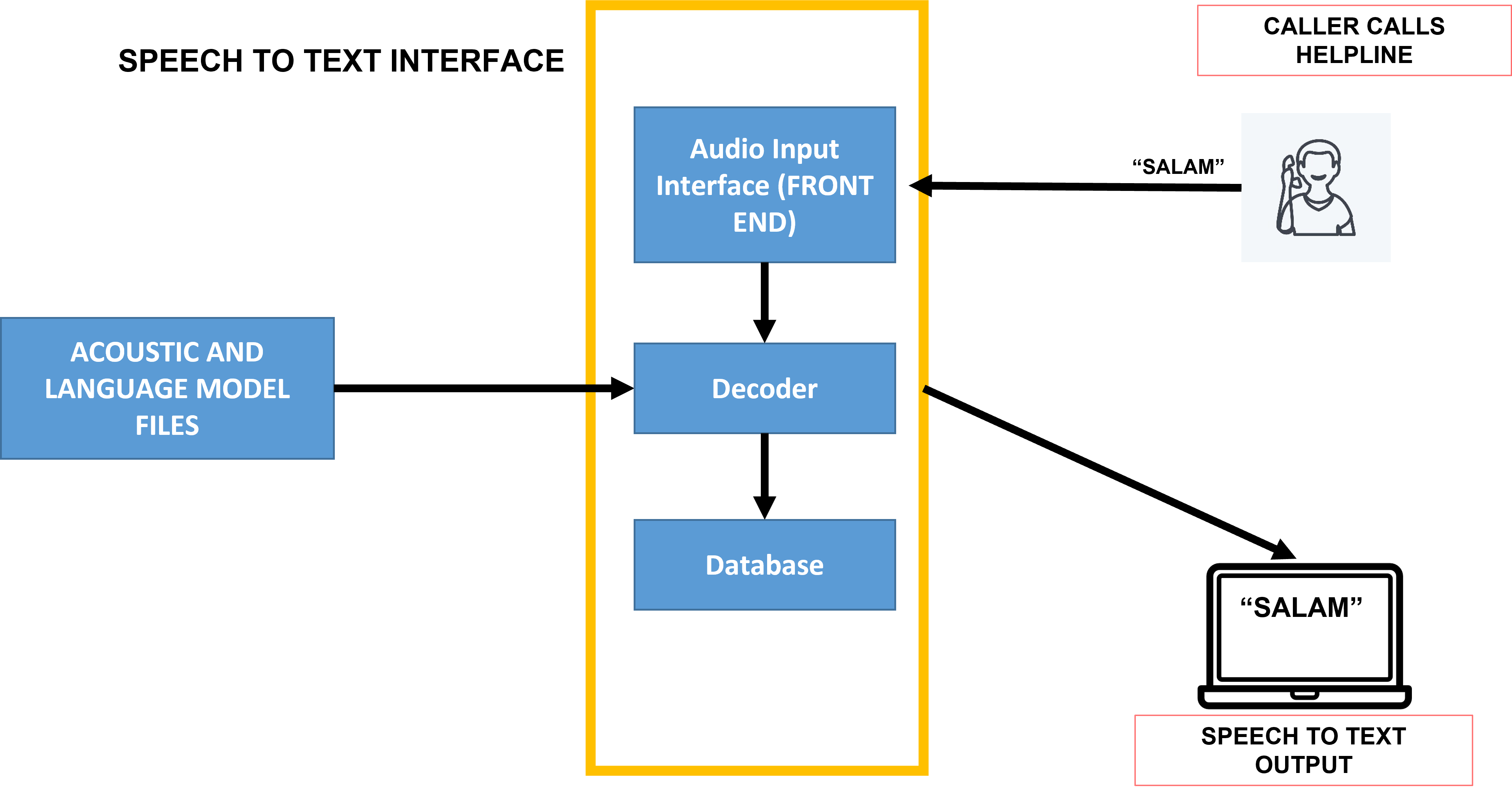}
    \caption{STT integration in Call Center}
    \label{fig:STT-working}
\end{figure}

ASR engines produce very huge model files which add to computational load, storage problems and cross platform deployment issues. We needed an interface that utilizes the necessary language and acoustic model files and provides Speech To Text services on audio signals (live or recorded),  reducing size of the model making it portable and cross-platform compatible. Working of STT in Call Center environment is shown in Figure \ref{fig:STT-working}.

We used an offline speech recognition library that comes with a set of accurate models, scripts, and practices and provides ready-to-use speech recognition for different platforms like mobile applications or Raspberry Pi since the model size reduced from 1-5GB to 50-500MB. We can also use this Platform to build a life-long learning platform \cite{eeckt_continual_2021} which continuously improves speech recognition for major languages and use cases \cite{alphacep_vosk_2022}.

This is the completion of our ASR implementation cycle and the Overall Workflow is summarized in Figure \ref{fig:workflow}.

\section{Experimental Setup} 
\label{sec:Implementation}

\subsection{Implementation System}
For implementation of our methodology, we used a Dell T7910, Dual Xeon Processors E5-2680 V4 @ 2.4GHz, RTX 3080Ti GPU, 32GB RAM, Ubuntu 20.04 Operating System. We installed the CUDA matrix library to access GPU-based matrix operations to be able to run required parts of GPU computation. We installed a GPU-accelerated primitives library for DNN called NVIDIA CuDNN which provides highly fine-tuned standard routines implementations like forward-backward convolution, pooling, normalization, and activation layers which gives effective performance with CNNs.

\subsection{ASR Engine}
\begin{figure*}[htbp]
    \includegraphics[width=0.95\textwidth]{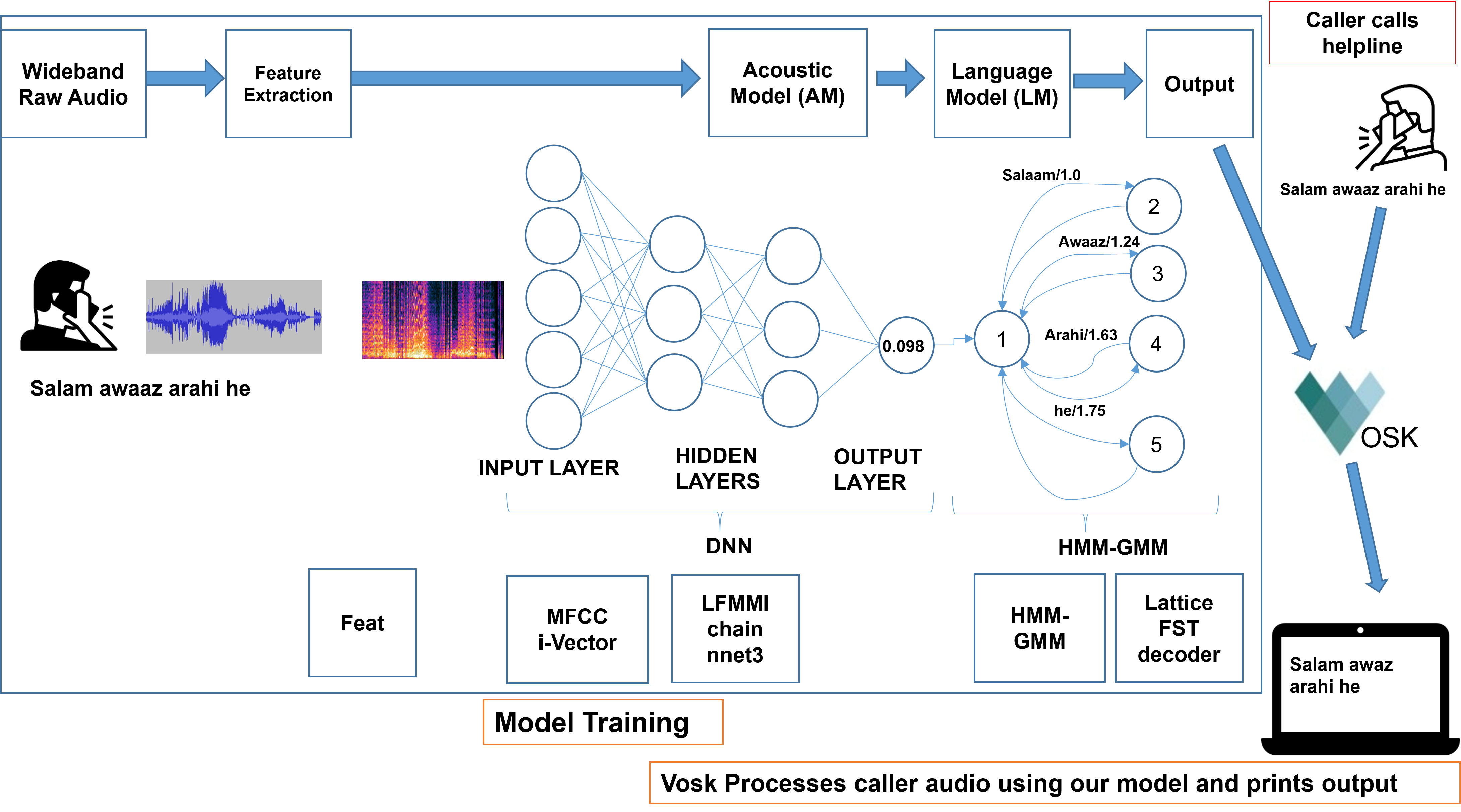}
    \caption{Implementation}
    \label{fig:implementation-architecture}
\end{figure*}

We used Kaldi ASR \cite{daniel_povey_kaldi_nodate} Toolkit for the preparation of Language and Acoustic Models and Training of ASR (using Chain nnet3 CNN-TDNN) as shown in Figure \ref{fig:implementation-architecture}. Kaldi is an open source research speech recognition toolkit that implements many state-of-the-art algorithms. Kaldi is currently one of the most tested and cited ASR engines with a very supportive open-source community dedicated to the development and expansion of the project. Various sources show that it still outperforms its contemporaries \cite{georgescu_performance_2021} \cite{christian_gaida_comparing_2014}.

\subsection{Dataset}
Dataset was divided into test and train and the Language Models were prepared using SRILM. Audio slicing and pre-processing were done using SoX, FFMPEG and Audacity. Monophone and Triphone HMM-GMM Training utilized Open FST, ATLAS and BLAS/LAPAC libraries where as in the Chain CNN-TDNN we used NVIDIA's CUDA and CuDNN for GPU Computation. The scripts used in this were mostly bash, C++ and Python.

Usually it is understood that more data means more accurate model but in any given scenario, Quality of data always trumps quantity, which is what we tried to ensure. A data-set with millions of rows with a lot of mislabelled observations can lead to an obscured learning process, while a smaller data-set with good quality labels tends to give much better results. The overall idea of taking a hybrid approach is to train a strong model with smaller dataset, use it to annotate the remaining unlabelled dataset. That data is then audited and improved over time which allows training of ASRs with other methods to give better robustness and accuracy.

With regard to the structured data-set the data organization was as follows:
\begin{enumerate}
    \item \textit{D1:} PRUS \cite{zia_pronouncur_2018} data set (along with \cite{qureshi_urdu_2021}) 
    \begin{itemize}
        \item Total 7 speakers, 708 sentence-utterances per speaker
        \item Train set = 4 and Test set = 3
    \end{itemize}
    \item \textit{D2:} Sehar Data-set \cite{sehar_gul_detecting_2020} 
    \begin{itemize}
        \item 100 Sentence-utterances (Malicious), 3 x speakers; 2 x Train and 1 x Test
        \item 86 Sentence-utterances, 7 x speakers; 5 x Train 2 x Test
        \item 5 Word-utterances, 13 x speakers; 10 x Train, 2 x Test
    \end{itemize}
    \item \textit{D3:} 250-word data set from Asadullah \cite{asadullah_automatic_2016} \cite{noauthor_urdu_nodate} which was also used by Sehar. This had the transcription in Urdu so we had to manually do the transcription in Roman Urdu since that was our chosen mode of writing to avoid spending time building different language models and blending them.    \begin{itemize}
        \item 10 Speakers;
        \item 8 x Train, 2 x Test
    \end{itemize}
    \item \textit{D4:} English and Urdu Alphabet data which we collected ourselves  
    \begin{itemize}
        \item 11 x Speakers in the Train set. Number of Utterances by each speaker was 28, 102, 206, 60, 28, 26, 37, 135, 256, 185 and 116 clips.
        \item 4 x Speakers in the Test set. Number of Utterances by each speaker was 52, 53, 44 and 55 clips 
    \end{itemize}
    \item \textit{D5:} Telephonic Data-set samples  
    \begin{itemize}
        \item Training - 5 x speakers and Utterances = 348, 8, 7, 55, 12 clips.
        \item For validation 38 utterances from 2 Speakers
    \end{itemize}
\end{enumerate}


\subsection{STT Interface}
ASR Model training produces a lot of Language and Acoustic Model Files which can be an issue during deployment and leads to cross-platform compatibility problems which is why we needed a Speech To Text (STT) Interface that could utilize the model files and give us flexibility and ease of deployment for which we used an open-source offline STT tool called Vosk \cite{alphacep_vosk_2022}. This required us to have python available since this portion is primarily python based.


\section{Experimental Results}
\label{sec:experimental-results}
We were able to achieve 13\% WER and SER of 31.2\% using HMM-GMM and with CNN-TDNN using LFMMI Objective Function we achieved 5.2\% WER and 20.45\% SER on clean and noisy code-switched Urdu audio data with read speech, spontaneous speech, isolated words and isolated digits. The details of our results is shown in Table \ref{tab:comparison-table}. 

For Evaluation we used WER and SER. While the option of using CER was there we found it irrelevant for us since we are more concerned with the accuracy of words and sentences. The Urdu word \textit{"Janbaaz"} can also be written as \textit{"Jaanbaaz"} or \textit{"Jaanbaz"} and will be spelled pretty much in the same way. Hence, as long as the overall word is accurate and the sentences are accurately transcribed, our ASR is suits our requirements.

It must be noted that Call center audio used for testing was totally unseen, containing OOVs, overlapping and low-volume speech. In audible speech signal areas with no overlap and clean speech (whether single word or long sentences) we were able to achieve WER 5.2\%. but in the overlapping areas, lower volumes, incomplete or partial speech by the speaker and OOVs, the performance degraded yielding 74\% result. 

\section{Analysis of Results}
\label{se:discussion}

The works we studied for Urdu ASR were tested to perform generally in clean environment on one of the following: Isolated words, Isolated Digits, Read speech or spontaneous speech. Our work was able to perform with 95\% accuracy on all these types of speech in a noisy as well as in clean environment. We found that most of the previous work generally focused on using pure Urdu words or sentences. The fact remains that Urdu is never spoken in pure form. It is always spoken in code-switch to English, Panjabi and other local languages as well in normal conversation. Moreover, from the perspective that we set out initially to improve work of \cite{sehar_gul_detecting_2020}, our model has clearly out-performed on Sehar's Data-set of malicious Urdu sentences and on general code-switched Urdu conversations as well. 


The usual idea for speech to text output may be to use local script for the transcription text and in Urdu it is usually Nastaliq. However, to address the code-switching issue we kept the written script as Roman instead of making Urdu Nastaliq and Roman English separately allowing us to bypass the requirement of building separate Language and Phonetic Models for every word of a different language used. Keeping one writing script for code-switched Low resource language will help solve the issue of Language modelling, saving a lot of time to structure and map data, while also reducing the Model complexity. 

Our Model was able to recognize numbers and write them down in Arabic Numeral scripts which was due to the fact that we mapped the numbers to their pronunciations in our Lexicon. In situations where the speech include series of number, the ASR used Arabic numeral as text output instead of their English pronunciation which is a result we have not seen before in a code-switched Urdu ASR. 


Our work also is a demonstration of effectiveness of Data Centric Approach in the long run. We built our ASR from scratch which meant that Data Pre-Processing also has to be iteratively audited along with model tweaking. This is one of the key reasons why our model performed better because errors were usually identified at the start rather than in a stage where debugging would be an ardous task. 

This approach improved accuracy by effectively using data as a strategic asset to ensure more precise estimates, observations, and decisions. It eliminates Complex data transformations and reduces Data inconsistencies, redundancy and errors. It improves data quality and reliability and reduces expenses since less time and effort is required to work on model building, tweaking and retraining.


We added a unique diversity in our data-set which covered maximum possible scenarios in our call center ASR. Our data performed optimally which includes (English and Urdu) Isolated Digits, Isolated words, Urdu Malicious sentences (Section \ref{sub:datasources} from D2), Read Speech (Words and Digits) of duration up to 20 seconds, Spontaneous Speech (Words and Digits) of duration up to 20 seconds, Clean Audio, Noisy or Telephonic Audio. Other models we studied in our literature review performed in on or two of the above points whereas our model covers all these scenarios and works as a speaker independent model. 

The best approach for a resource constraint scenario is to take small data-set of 10-20 hours. We accurately label it to build a strong language model. It is best to understand the context of data-set and do an analysis of the language usage e.g. for code-switching, common words, jargon etc. Then train a strong model which can then annotate the remaining data-set which can then be used to train Deep Learning Model which is known to improve as data-set size increases. The golden data-set we will eventually have as we accurately label and integrate more and more data will become our asset. Models will evolve for the better but data-set is the same (mostly a standard). Hence it is always best to build up the data-set instead focusing on model only.

HMM-DNN method involves use of statistical model to build up Language and acoustic models. These alignments are used to further train and improve model using Neural Networks. In a way it uses the accuracy of Statistical Models on low data-sets and the scalability of Neural Networks together. If the input features are noisy, having a few CNN layers (2D CNN) before TDNN (which is basically a time dilated 1-dimension CNN) blocks at the input would help the model learn some non-linear transformations which can reduce the effect of the noise on speech input. TDNNs are basically temporal convolution whereas CNNs are convolutions in both frequency and space dimensions, which gives it the capability to learn patterns better. Speech Recognition requires a system that takes care of temporal context and time delay neural network architecture captures long term temporal dependencies better than a feed-forward DNN does. We need to look at features left and right i.e. contexting and we need better prediction of phoneme states which is why for Low Resource Languages TDNNs work best. 

The LF-MMI or chain model helped reduce the training time of the model and allowed for more efficient and effective training which improved accuracy from HMM-GMM Triphone training by 8\% in our case. It makes best use of GPU and avoids using branching or pruning tree search. The process of creating the denominator FST is very similar to that of a traditional decoding graph creation, however the computation of the denominator forward-backward computation are parallelized by the GPU. The process is sped up by performing careful optimizations, including reversal, weight pushing, and minimization followed by epsilon removal on the denominator FST to minimize graph size. Hence this reduces graph size and improves training speed an performance while using Neural Networks, in comparison to pure Deep Learning based models which is why it is also applicable on streaming data.


We identified based on relevant work from our literature review that Model Files that are generated are often huge in Size which causes difficulty in deployment. Our proposed framework utilized a Speech To Text Interface that would use relevant files giving it flexibility, ease of cross-platform deployment and scalability \cite{alphacep_vosk_2022}. Using Kaldi and Vosk \cite{alphacep_alphacepvosk-server_2022} allows us to built accurate offline speech recognition supporting major communications protocols like MQTT, GRPC, WebRTC, and Websocket. Server can be locally used to provide speech recognition to smart homes and PBXs such as free-switch or asterisk and can also deployed as a back-end for web-based streaming speech recognition, as well as to power chat-bots, websites, and telephony.

\section{Comparison with Relevant Work} 
\label{sec:comparison_result}

Since, We lacked access to free or open-source corpus resources that others could adopt and use for a fair comparison which is why we used data-set from Ali et al 2015 \cite{ali_automatic_2015} and PRUS (2018) \cite{zia_pronouncur_2018, qureshi_urdu_2021} results of which were compared with our model in Section \ref{sec:comparison_result}.

Moreover, we used only WER as evaluation metric since we only had WER available in those work as shown in Table \ref{tab:comparison-table}. Our Call Center data is not openly available and hence cannot be used as a metric to compare performance with others. However we can use \cite{ali_automatic_2015}, \cite{sehar_gul_detecting_2020} and \cite{qureshi_urdu_2021} for bench-marking since their data was available and applied in works other than ours. 


All these works focused on one of the aspects of ASR input i.e. read speech, isolated words or digits and spontaneous speech. In our case all of these scenarios were covered which makes it more applicable in practical sense. The works in comparison were all tested primarily on clean audio, making it unsuitable for noisy telephonic environment. In our case the model performed with 5.2\% WER in noisy and clean environment. However in overlapped, improper pronunciations and low volume speech, our model was unable to perform well.

These works focused on pure Urdu words which means they are not applicable for telephonic environment since languages are spoken in code-switched manner. We were able to identify words usage patterns based on Call Center scenario to cover maximum possible vocabulary, trained with noisy and clean samples of same word set to cover maximum possible scenario. We also solved the code-switching problem by keeping a unified written script i.e. Roman. So all English or Urdu words were in Roman script which eased the lexicon building and language modelling process. 

One of the drawbacks of \cite{sehar_gul_detecting_2020} that we identified was that it used very little data-set to train Urdu ASR with Deep Learning. In our case we had limited access to labelled dataset and most of our dataset was unlabelled for which we trained a Hybrid HMM-DNN ASR which was able to use statistical alignments and Neural Networks to train accurate ASR on limited dataset. Hence our approach is suitable particularly for Low-Resource Languages where labelled dataset is not easily available.

Table \ref{tab:comparison-table-other} shows comparison with other relevant recent work in ASR but the comparison does not apply to our scenario because our problem area is code-switched Urdu in noisy telephonic environment even if the results are better than our work.

\begin{table}[h]
\centering
\caption{Comparison of our Results with relevant work}
\label{tab:comparison-table-other}
\resizebox{\columnwidth}{!}{%
\begin{tabular}{|l|l|l|l|c|}
\hline
Author & Dataset & Method & WER & \multicolumn{1}{l|}{SER} \\ \hline
Our Work & Data-set \ref{sub:datasources} D(1-5) & HMM GMM TRIPHONE & 13\% & \multicolumn{1}{l|}{31.2\%} \\ \hline
Our Work & Data-set \ref{sub:datasources} D(1-5) & LFMMI CNN-TDNN & 5.2\% & \multicolumn{1}{l|}{20.45\%} \\ \hline
\multirow{2}{*}{Georgescu (2021) \cite{georgescu_performance_2021}} & \multirow{2}{*}{LibriSpeech English Dataset (other/noisy)} & CNN-TDNN & 3.85\% & - \\ \cline{3-5} 
 &  & TDNN & 3.87\% & - \\ \hline
\multirow{2}{*}{Alsayadi (2021) \cite{alsayadi_arabic_2021}} & \multirow{2}{*}{1200 hours Arabic Corpus} & Conventional (HMM-DNN) & 33.72\% & - \\ \cline{3-5} 
 &  & CNN-LSTM (E2E) & 28.5\% & - \\ \hline
Lakshmi (2020) \cite{lakshmi_sri_kaldi_2020} & 158 Hindi medical terms, 27 speakers (16 female and 11 male), 3 hours, 16KHz, mono-channel & Triphone HMM-GMM & 2\% & - \\ \hline
\end{tabular}%
}
\end{table}

Table \ref{tab:comparison-table} shows comparison with relevant Urdu works that we could find. Datasets from these works were also used which allowed us to have a fairer comparison.


\begin{table}[h]
\centering
\caption{Comparison of our Results with relevant work in Urdu}
\label{tab:comparison-table}
\resizebox{\columnwidth}{!}{%
\begin{tabular}{|l|l|l|l|l|}
\hline
Author                                                               & Dataset                                                                                  & Method           & WER     & SER                    \\ \hline
Our Work                                                             & Data-set \ref{sub:datasources} D(1-5)                                   & HMM GMM TRIPHONE & 13\%    & 31.2\%                 \\ \hline
Our Work                                                             & Data-set \ref{sub:datasources} D(1-5)                                   & LFMMI CNN-TDNN   & 5.2\%   & 20.45\%                \\ \hline
Qasim \cite{qasim_urdu_2016}                      & 139 District names of Pakistan, 300 speakers                                             & HMM              & 24.5\%  & \multicolumn{1}{c|}{-} \\ \hline
Hazrat Ali \cite{ali_automatic_2015}              & 205 isolated words (20 Speakers)                                                         & HMM              & 33\%    & \multicolumn{1}{c|}{-} \\ \hline
Zoraiz Qureshi \cite{qureshi_urdu_2021}           & PRUS\cite{zia_pronouncur_2018} 708 sentences data set with 5 speakers & GMM/HMM          & 25-35\% & \multicolumn{1}{c|}{-} \\ \hline
\multirow{5}{*}{Naeem \cite{naeem_subspace_2020}} & \multirow{5}{*}{Dataset by RUMI and CSALT and NUCES - clean Mono-16000Hz}                & MONO             & 32.4\%  & 64.26\%                \\ \cline{3-5} 
                                                                     &                                                                                          & Tri-1            & 16.70\% & 52.41\%                \\ \cline{3-5} 
                                                                     &                                                                                          & Tri-2            & 16.74\% & 52.80\%                \\ \cline{3-5} 
                                                                     &                                                                                          & Tri-3            & 13.63\% & 52.62\%                \\ \cline{3-5} 
                                                                     &                                                                                          & HMM-GMM-SGMM     & 9.64\%  & 48.53\%                \\ \hline
\end{tabular}%
}
\end{table}

\section{Conclusion} 
\label{cha:discussion_conclusion}
Our work presented a data-centric approach to train a code-switched Urdu ASR using Hybrid HMM-DNN method for noisy telephonic environment in a resource-constrained scenario. We analyzed the call center call content and built our dataset. To cater for the code-switching issue between Urdu and English we used Roman script for Text output. The from various sources mentioned in section \ref{sec:Data_preprocessing} were collected, labelled and processed. We trained a HMM based model followed by Chain CNN-TDNN for model building. This was then linked to a Speech To Text Interface allowing the model to be flexibly used across platforms.


A robust speech recognition system generally requires hundreds of hours of transcribed speech data, as well as a massive text corpus and lexicon which is lacking with under-resourced languages like Urdu which is why Urdu speakers are around the world excluded from the benefits of ASR.

We proposed a workable framework that involved generation of statistical modelling and alignment for language and acoustics of speech data and applied Neural Networks to improve the accuracy of ASR system. With only 10 hours of data we were able to achieve up to 5.2\% WER. This results shows that Hybrid HMM-DNN (Chain CNN-TDNN) method is an effective and efficient method for ASR Model training with limited data-set available. 

However the model alone is not the only thing to consider when working on a deployable solution. Our work is one of the few examples that takes a Data-Centric Approach to machine learning and the primary reason why our framework gave good results was due to our focus on improving data before tweaking the model.

We had initially set out, only to improve model of \cite{sehar_gul_detecting_2020} to detect malicious words but we ended up building a solution which not only outperformed on data set from \cite{sehar_gul_detecting_2020,ali_automatic_2015,qureshi_urdu_2021}, but we also managed to improve its performance for read-speech, continuous or spontaneous speech and isolated word utterances in code-switched Urdu environment. Our work however still requires improvement in it's robustness especially in overlapping speech signals. Our work also took into account various security considerations. In fact security policies and practices with respect to ASR systems is definitely and important research area which can be explored further in future. We integrated an open source Speech To Text Interface that allows the models to be easily deployed in Desktop, IOT devices, Asterisk (open source call center solution) Web and Client-Server Environment. 

Hence the work paves a way ahead for practical use of Code-Switched Urdu ASR in a noisy telephonic environment to benefit Urdu Speaking people around the world.

\section{Acknowledgments}
We thank NUST-PNEC for supporting our work. We thank Citizen Police Liaison Committee for supporting the practical implementation of our research work. We are also grateful to all our colleagues and friends from open-source communities who helped us with their valuable suggestions that helped improve our work.

\newpage

\section*{Biography}
This sections covers the biography of the authors of this work:
 
\begin{figure*}[h]
\includegraphics[width=1in,height=1.5in,clip,keepaspectratio]{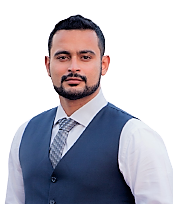}
\textbf{Muhammad Danyal Khan} is a Cyber Security Professional who did his BS Management Information System \& MS in Cyber Security from National University of Science \& Technology (NUST-PNEC) during course of his Masters, he helped CPLC implement of Speech To Text Systems for Call Center. He has worked in establishment of Basic and Advanced CyS Lab, as faculty member and Head of Post Graduate Program of CyS Department in PNEC-NUST. He has also served as Operations Branch Naval Officer for 12 years and has experience in various fields of Naval Operations like Above/Surface/Under Water Warfare, amphibious operations, close quarter combat, Navigation, Field Intel-gathering, ELINT, COMINT, SIGINT, Cyber Warfare and Information warfare. His areas of interest include Network Security, Information Warfare, ASR, IOT, DevSecOps, Edge AI, Secure Systems Architecture \& Secure MLOPS.
\end{figure*}

\begin{figure*}[h]
\includegraphics[width=1in,height=1.5in,clip,keepaspectratio]{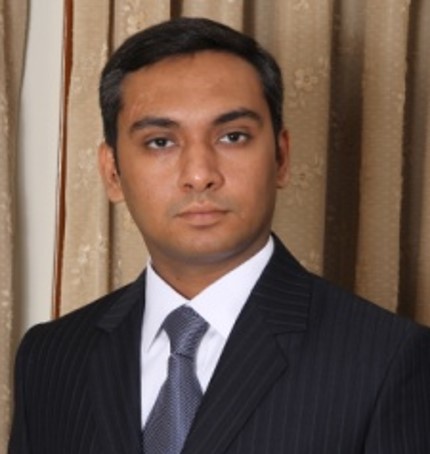}
\textbf{Raheem Ali} is an experienced Software Architect who has worked for various banking, corporate and government organizations in the field of DevOps and Cyber Security. His major contribution to the field of Encryption is the invention of Elliptic Chaotic Cryptography. He has been working on development of Linux Kernel for more than a decade and has also worked for various open source projects.
\end{figure*}

\begin{figure*}[h]
\includegraphics[width=1in,height=1.5in,clip,keepaspectratio]{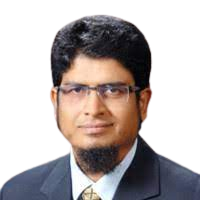}
\textbf{Professor Dr. Arshad Aziz} is an experienced Professor with a demonstrated history of working in the higher education industry. Skilled in Cyber Security, Field-Programmable Gate Arrays (FPGA), Voice over IP (VoIP), Internet of Things and Software Defined Networking. He has a strong education professional with a Doctor of Philosophy - PhD (EE) focused in Cyber Security, FPGA Based System Design from National University of Sciences and Technology (NUST).
\end{figure*}
\vfill

\clearpage \newpage

\bibliographystyle{ieeetr}
\bibliography{references}  






\end{document}